\documentclass[11pt]{article}

\usepackage[final]{acl}

\usepackage{times}
\usepackage{latexsym}
\usepackage{amsmath}
\usepackage{amssymb} 
\usepackage{fontawesome}
\usepackage{float}

\usepackage{algorithm}
\usepackage{algpseudocode}

\usepackage{todonotes}
\usepackage{booktabs} 

\usepackage{multirow}
\usepackage{multicol}
\usepackage{tabularx}
\usepackage{longtable}
\usepackage{wrapfig}
\usepackage[breakable]{tcolorbox}
\usepackage{subcaption}

\usepackage[T1]{fontenc}

\usepackage[utf8]{inputenc}

\usepackage{microtype}

\usepackage{inconsolata}

\usepackage{graphicx}
\usepackage{adjustbox}
\usepackage[table]{xcolor}

\usepackage{soul}
\DeclareRobustCommand{\good}[1]{{\sethlcolor{green!10}\hl{#1}}}
\DeclareRobustCommand{\bad}[1]{{\sethlcolor{red!10}\hl{#1}}}

\DeclareRobustCommand{\styleLow}[1]{{\sethlcolor{blue!10}\hl{#1}}}
\DeclareRobustCommand{\styleHigh}[1]{{\sethlcolor{orange!20}\hl{#1}}}

\DeclareRobustCommand{\styleLowMax}[1]{{\sethlcolor{blue!25}\hl{#1}}}
\DeclareRobustCommand{\styleLowMid}[1]{{\sethlcolor{blue!10}\hl{#1}}}
\DeclareRobustCommand{\styleHighMid}[1]{{\sethlcolor{orange!15}\hl{#1}}}
\DeclareRobustCommand{\styleHighMax}[1]{{\sethlcolor{orange!35}\hl{#1}}}

\newcolumntype{L}{>{\raggedright\arraybackslash}X}

\title{\textsc{LiteraryBigFive}: Author-Personalized Text Generation \\ in a Unified Interpretable Space}

\author{
 \textbf{Jinghui Zhang\textsuperscript{1}},
 \textbf{Lang Gao\textsuperscript{1}},
 \textbf{Ao Li\textsuperscript{2}},
 \textbf{Mingzhe Li\textsuperscript{3}},\\
 \textbf{Ruihong Zeng\textsuperscript{1}}, 
 \textbf{Zirui Song\textsuperscript{1}},
 \textbf{Kentaro Inui\textsuperscript{1,4,5}},
 \textbf{Xiuying Chen\textsuperscript{1,*}}
\\
 \textsuperscript{1}MBZUAI,
 \textsuperscript{2}Shandong University,
 \textsuperscript{3}Independent Researcher,
 \textsuperscript{4}Tohoku University,
 \textsuperscript{5}RIKEN
\\
 \small{
   \texttt{\{jinghui.zhang,lang.gao,ruihong.zeng,zirui.song,kentaro.inui,xiuying.chen\}@mbzuai.ac.ae} 
   } \\
   \small{
    \texttt{liaolea@mail.sdu.edu.cn, li\_mingzhe@pku.edu.cn}
    }
}

\begin{document}
\maketitle
\begingroup
\renewcommand{\thefootnote}{}
\footnotetext{* Corresponding author.}
\addtocounter{footnote}{-1}
\endgroup

\begin{abstract}
Personalized text generation for authors and literary writing is essential for applications such as adaptive writing assistants, creative support tools, and computational literary analysis.
However, existing approaches to author modeling and personalization often represent writing behavior as \textit{independent labels}, requiring large-scale corpus collection or fine-tuning for each author or stylistic category.
Such formulations are costly, difficult to interpret, and poorly suited for generalizing across authors.
Inspired by the Big Five model's dimensional view of personality, we propose \textsc{LiteraryBigFive}, a framework that reframes authorial writing characteristics as coordinates within a \textit{unified and interpretable space.}
In this space, we derive each interpretable axis (e.g., Classicism, Emotionality) from activation-space contrasts between author-written and neutral passages, yielding distinct stylistic dimensions that allow texts or authors to be positioned within a five-dimensional system.
Beyond localizing different authors, we further introduce an interpretable steering mechanism, which adaptively guides text generation toward target coordinates to perform author-personalized writing.
Experimental results show that \textsc{LiteraryBigFive} improves authorial expressiveness while preserving semantic fidelity. 
The derived author per-axis scores strongly correlate with real-world literary consensus, offering transparent and interpretable explanations of author-specific generation behavior:
\faGithub~\href{https://github.com/Znull-1220/LiteraryBigFive}{Github}.
\end{abstract}

\section{Introduction}

Personalized text generation models individual writing styles, especially for authors and literary writers with distinctive voices. 
It supports stylistic preference matching and voice emulation in applications such as creative writing~\cite{yu2024charpoet,qin2025writingsupport} and personalized assistants~\cite{zhang2025fromindividual,lin2025userllm}.

\begin{figure}[t]
\centering
  \includegraphics[width=\columnwidth]{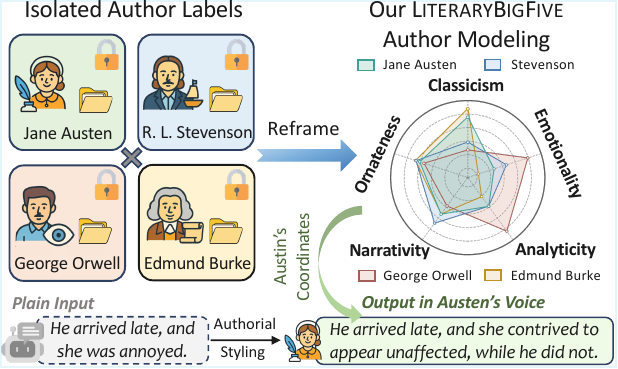}  
  \caption{Previous work models each author as an isolated label; \textsc{LiteraryBigFive} reframes authorial characteristics as a unified, interpretable space spanned by five axes, enabling measurement, comparison, and control across authors and books.}
  \label{fig:intro}
  \vspace{-2mm}
\end{figure}

However, existing approaches predominantly treat individual authors as isolated categories.
As shown in Figure~\ref{fig:intro}, previous methods typically set up a separate task for each author, and apply specific prompting~\citep{bhandarkar2024emulating}, model training~\citep{jhamtani2017shakespearizing}, or steering~\citep{konen2024style} to capture their characteristics independently.
This design suffers from two key limitations. 
First, adapting to a new author typically requires collecting hundreds or thousands of texts or retraining the model~\cite{zhang2025personalization}, making large-scale expansion extremely costly and impractical.
Second, it fails to reveal how different writing patterns relate to one another, offering limited interpretability or a unified representation of authorial variation.

Linguistic and literary studies offer a more systematic and theoretically grounded perspective for understanding complex writing patterns and stylistic variation across texts.
Decades of analysis show that variation in written language is often organized along a few stable and interpretable dimensions, such as narrativity, emotion, and elaboration, rather than an unlimited and highly fragmented set of individual author labels~\cite{martin2003language,kuiken2021handbook,biber2016grammaticalcomplexity}.
This dimensional view echoes the idea behind the Big Five model in psychology, 
where complex human personalities are described by five high-level axes~\cite{goldberg1993structure,john1999bigfive}.
These works suggest that \emph{unified, interpretable coordinates} could support a more flexible paradigm for author personalization than categorical tags.

Building on these observations, we propose \textsc{LiteraryBigFive}, a framework that reframes authorial characteristics as coordinates in a unified five-dimensional space rather than a set of unrelated labels.
Concretely, we define five interpretable axes: Classicism, Ornateness, Narrativity, Emotionality, and Analyticity for our \textsc{LiteraryBigFive}, 
which are informed by established literary and linguistic analysis~\citep{BiberConrad2019Register,abbott2021narrative,biber2016grammaticalcomplexity,booth1983rhetoric}.
To construct the space, we select representative classics for each dimension and derive axis directions by contrasting original author-written and neutral passage pairs that preserve semantics while varying axis-specific features.
Since raw activations show a general shift from neutral rewrites toward original literary texts that entangle distinct axes, we introduce an axis decomposition step to explicitly remove the shared principal component from all axes and reinterpret it as the overall expressiveness direction, thereby yielding the refined BigFive axes system.
This improves axis independence and enables more stable multi-axis control over individual authorial traits.

With the \textsc{LiteraryBigFive} space established, we propose \emph{localize-and-steer} for both authorial coordinates analysis and personalized generation.
For a target author, we first \textit{locate} their position by projecting the reference text onto the BigFive directions, and yield unique scores that capture their authorial characteristics.
This allows us to position different authors or books within a unified space and compare them along shared dimensions.
Next, we leverage the BigFive axes for interpretable personalized generation by \textit{steering} model activations toward target authorial coordinates, enabling immediate adaptation to a wide range of new authors, from classic novelists to contemporary writers.

To validate the effectiveness of \textsc{LiteraryBigFive} on unseen authors, we evaluate it on books spanning distinct writing identities.
Experiments show that \textsc{LiteraryBigFive} better matches target authors while preserving meaning.
Meanwhile, the learned author coordinates align with established literary consensus, suggesting that the space provides an interpretable representation of style.

In general, our contributions can be summarized as follows:
(i) We introduce \textsc{LiteraryBigFive}, a framework motivated by linguistic and literary studies that advances author modeling from isolated labels to a unified, interpretable five-dimensional space, capturing core dimensions of authorial variation.
(ii) We propose a localize-and-steer mechanism that maps individual authors to precise coordinates in the \textsc{LiteraryBigFive} space and enables latent space steering for personalized generation, adapting to new authors instantly without retraining.
(iii) We demonstrate that \textsc{LiteraryBigFive} improves authorial expressiveness while preserving semantic fidelity, and that the resulting axis scores align well with established literary consensus, providing transparent, per-dimension explanations of authorial characteristics.

\section{Related Work}

\textbf{Personalized Text Generation.}
Personalized generation aims to align LLMs with specific user profiles or authorial identities while preserving semantic content~\cite{zhang2025personalization}.
Traditional approaches often frame this as a supervised rewriting task requiring parallel corpora~\cite{hu2017toward}, or employ unsupervised disentanglement to separate content from linguistic expression~\cite{prabhumoye2018style}.
In the era of LLMs, the research focus has shifted to prompting~\cite{reif2022recipe} or fine-tuning on author-specific corpora~\cite{wang2024rolellm}.
However, these methods typically treat individual authors as \textit{independent, categorical labels}.
This label-based paradigm scales poorly, as modeling a new author needs separate corpus collection, modeling retraining or extensive prompt engineering. 
We address this by learning a unified latent space that adapts to new authors instantly without such per-author overhead.


\noindent\textbf{Dimensional Modeling of Linguistic Variation.}
Traditional stylometry often treats authors discretely, assigning each author a unique label and modeling style differences as class distinctions~\cite{holmes1998evolution}.
In contrast, linguistic studies have shown that written language can also be characterized along multiple interpretable dimensions, such as \textit{involved} versus \textit{informational} writing~\cite{biber1991variation,biber2016grammaticalcomplexity}.
These studies provide an empirical basis for dimensional analysis of linguistic variation, but are not designed for controlling language model generation at inference time.
Beyond linguistics, the Big Five framework in psychology also shows that complex human individual variation can be described through compact interpretable axes~\cite{goldberg1993structure}, and this dimensional view has been widely adopted in NLP research for assessing personality~\cite{jiang2024personallm} and simulating personas~\cite{wang2024rolellm}.
However, it remains underexplored for controllable personalized text generation, where categorical style or author labels still dominate.
Our work builds on this dimensional perspective and represents authorial writing characteristics as coordinates in a shared, interpretable space, enabling both author localization and activation-based steering for personalized generation.

\noindent\textbf{Activation Steering.}
Activation steering modifies a model’s output at inference time by intervening in its intermediate representations using direction vectors ~\cite{zou2023transparency}.  
By computing difference in latent activations between samples that express a target concept and those that do not, one can isolate a semantic vector corresponding to a specific attribute, and steering the model along this direction induces the associated behavior~\cite{KimWGCWVS18}.  
Key advantages of activation steering include its interpretability, as abstract attributes are explicitly represented as vectors~\cite{rimsky2024steering,gao2026the}, as well as its efficiency compared to conventional adaptation methods such as finetuning~\cite{abs-2508-17621}.
Recent studies have applied activation steering to personalized writing~\cite{zhang2025personalized}, emotion control~\cite{abs-2510-04484}, and persona adoption~\cite{abs-2507-21509}.  
Although existing methods can steer models toward target outputs, they are mostly limited to single, binary traits or learn distinct vectors for individual authors.
In contrast, \textsc{LiteraryBigFive} enables multi-dimensional personalized steering within a unified space across diverse authors.

\section{Problem Formulation}

We formulate interpretable author-personalized generation as two coupled sub-tasks: \textit{localization} and \textit{steering}, jointly framed in a unified five-dimensional space.
In the localization stage, let $\mathcal{S}=\mathbb{R}^5$ denote the \textsc{LiteraryBigFive} space with interpretable axes.
Given few $k$ reference passages $\{x_{b,i}\}_{i=1}^k$ sampled from a target book $b$, 
a locator $\phi:\mathcal{T}\!\to\!\mathcal{S}$ maps each passage to its coordinates in $\mathcal{S}$.
We estimate the target authorial coordinates by averaging the coordinates of its passages:
\begin{align}
    \mathbf{s}_b=\textstyle \frac{1}{k}\sum_{i=1}^k \phi(x_{b,i}) \in \mathcal{S},
    \label{phi}
\end{align}
which represents the author’s or book’s characteristic position within the space.

In the steering stage, given a neutral input passage $x$ and the target position $\mathbf{s}_b$, 
the goal is to generate a rewritten passage:
\begin{align}
    \hat{x} = f(x, \mathbf{s}_b),
\end{align}
whose semantics remain consistent with $x$ while its linguistic expression aligns with $\mathbf{s}_b$.

\section{Method}
In this section, we introduce the \textsc{LiteraryBigFive} in detail.
First, \S \ref{space} presents the space construction with five axes;
then, \S\ref{localization} describes how to locate a book to obtain its authorial coordinates;
and finally, \S\ref{steer} explains how to steer generation to the target author using these coordinates.

\begin{figure*}[t]
  \includegraphics[width=\linewidth]{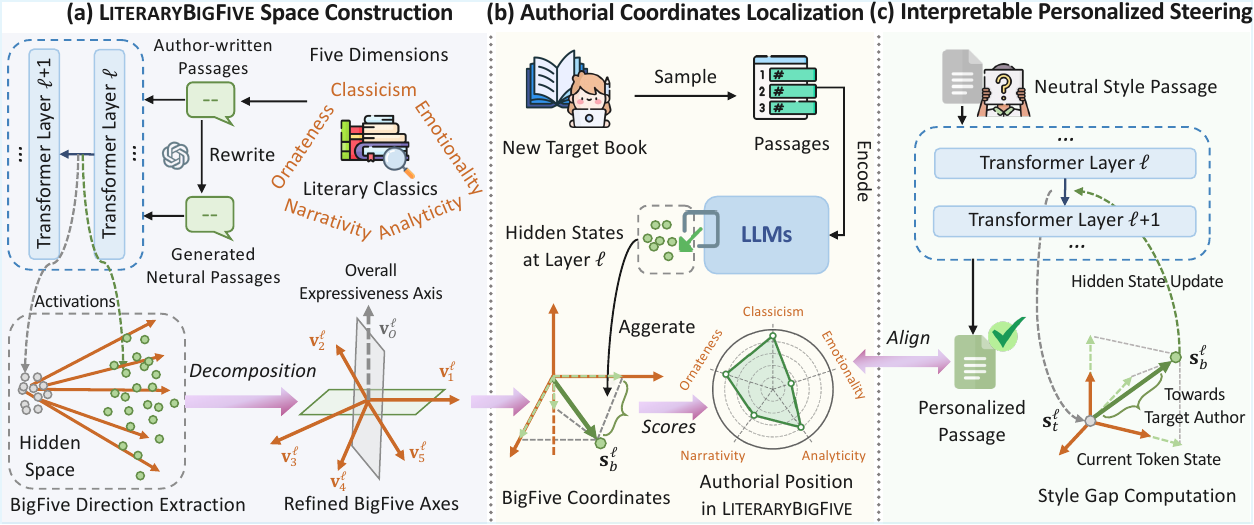} 
  \caption {Illustration of \textsc{LiteraryBigFive} framework. (a) We begin with constructing the \textsc{LiteraryBigFive} space using author-written passages from selected literary classics that strongly exhibit each defined dimension, and extract BigFive direction with \textit{axis decomposition}. 
 (b) For a new target book, \textsc{LiteraryBigFive} \textit{locates} its authorial position by projecting reference passages onto the BigFive axes to obtain authorial coordinates. 
  (c) During generation, we align the output with the target author by computing the \textit{style gap} between the current token and the target coordinates, then updating the hidden states along the interpretable axes to close this gap.}
  \label{fig:method}
  \vspace{-5mm}
\end{figure*}

\subsection{\textsc{LiteraryBigFive} Space Construction}
\label{space}
\paragraph{Dimension Definitions and Representative Books.}
Drawing on prior work in linguistic and literary analysis \cite{BiberConrad2019Register},
we define five interpretable dimensions to capture several major variations in English literary writing.
Ornateness reflects lexical richness and syntactic complexity, particularly within noun phrases, rather than simple sentence length \cite{biber2016grammaticalcomplexity}.
Narrativity distinguishes storytelling text, focused on action verbs and time markers~\cite{abbott2021narrative}.
Emotionality quantifies affective intensity through sentiment words, regardless of the specific topic \cite{booth1983rhetoric}.
Classicism reflects the traditional writing patterns of the 18th and 19th centuries, emphasizing complex sentence structures and historical vocabulary \cite{boyd2022development}.
Analyticity corresponds to expository and reasoning-oriented writing, characterized by abstract nouns and logical relations \cite{biber1995dimensions}.

For each dimension, we select representative classics discussed in literary analysis to anchor the corresponding axis.
Examples include Daniel Defoe's Robinson Crusoe, a foundational work for modern linear Narrativity~\cite{watt1957rise}, and Virginia Woolf's Mrs. Dalloway, distinguished by the intense Emotionality of its affective experience~\cite{auerbach2013mimesis}.
We curate selected books and clean these texts to retain only the authorial content and segment them into coherent passages.
The full list of authors and books, detailed data sources, and preprocessing procedures are provided in Appendix~\ref{app:dataset}.

\paragraph{Paired Dataset Construction.}  

To derive the axes for the five dimensions, we construct paired passages that share semantics but differ in authorial expression.
For each dimension \(k \in \{1, \dots, 5\}\), we denote the set of representative books as \(\mathcal{B}_k\).  
Following \cite{ma2025dressing}, for each cleaned author-written passage \(x_{b,i}^{+}\) from a book \(b \in \mathcal{B}_k\), we use GPT-4~\cite{openai2024gpt4technicalreport} to suppress authorial cues along the five defined dimensions (prompt in Appendix~\ref{app:neutralization}), obtaining a semantics-preserving neutral rewrite $x_{b,i}^{-}$, yielding $N_b$ passage pairs:
\begin{equation}
    \mathcal{P}_{b,k} = \{(x_{b,i}^{+}, x_{b,i}^{-})\}_{i=1}^{N_b}.
\end{equation}
The collection of all pairs for dimension \(k\) forms the dataset \(\mathcal{D}_k = \textstyle \bigcup_{b \in \mathcal{B}_k} \mathcal{P}_{b,k}\),  
and the complete corpus for axis construction is \(\mathcal{D} = \bigcup_{k=1}^{5} \mathcal{D}_k\).

\paragraph{Axis Extraction.}
After defining the five dimensions and preparing representative author data for each axis, we next extract the corresponding \textit{axis directions} in hidden space.
Let $a^{\ell}(s)\in\mathbb{R}^d$ denote the last-token activation at layer $\ell$ for a token sequence $s$. 
The key is to identify, for each dimension, the activation shift that captures authorial variation rather than semantic content or positional bias. 
Hence, for each paired passage $(x^+_{b,i},x^-_{b,i})$, 
we use the same neutral input $x^-_{b,i}$ and concatenate it with either the author-written passage $x^+_{b,i}$ or the neutral rewrite $x^-_{b,i}$, and then compute the \emph{author-written} and \emph{neutral} hidden states:
\begin{equation*}
\mathbf{h}^{\ell}_{+,b,i}=a^{\ell}(x^{-}_{b,i}\oplus x^{+}_{b,i}),\quad
\mathbf{h}^{\ell}_{-,b,i}=a^{\ell}(x^{-}_{b,i}\oplus x^{-}_{b,i}),
\end{equation*}
where $\oplus$ concatenates the neutral input and the model’s rewritten output. 
Since the input $x^-_{b,i}$ is identical in both sequences and only the output’s \emph{style} differs, we can obtain the  contrast $\boldsymbol{\delta}^{\ell}_{b,i} =\mathbf{h}^{\ell}_{+,b,i}-\mathbf{h}^{\ell}_{-,b,i}$ that isolates stylistic differences.
Finally, we average these vectors over all $N$ pairs from anchor books and renormalize, yielding the raw axis $\tilde{\mathbf{v}}^{\ell}_{k}$ for the $k$-th dimension at layer~$\ell$:
\begin{equation} \tilde{\mathbf{v}}^{\ell}_{k}=\textstyle \frac{1}{N}\sum_{i=1}^{N}\boldsymbol{\delta}^{\ell}_{b,i}\in\mathbb{R}^{d}. 
\end{equation}

\begin{figure}[t]
\centering
\includegraphics[width=\columnwidth]{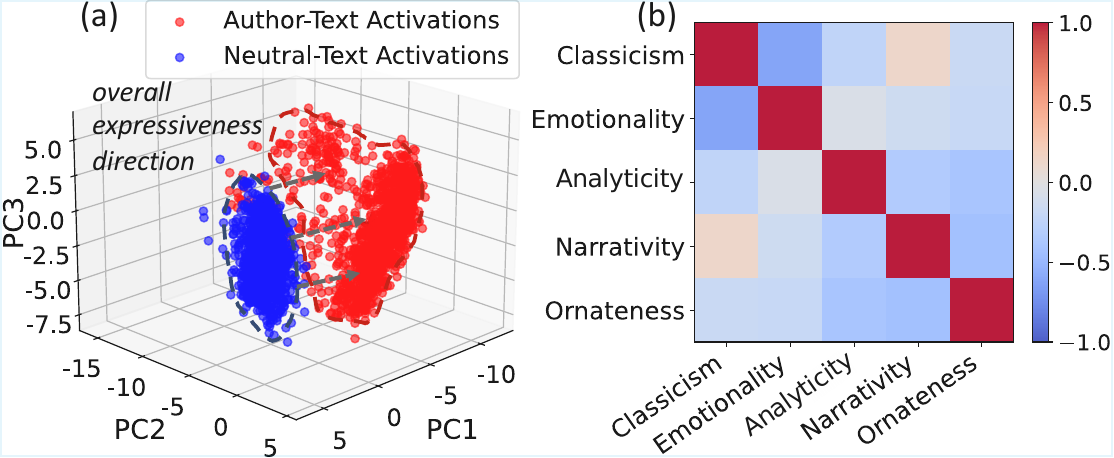}  
\caption{
(a) 3D PCA reveals a dominant expressiveness direction from neutral-text (blue) to author-text (red) activations. 
(b) Correlation heatmap shows reduced cross-axis similarity after removing this global trend, demonstrating effective disentanglement of our refined dimensions. 
Results shown for LLaMA2-7B-Chat.
}
  \label{fig:axis_analysis}
    \vspace{-2mm}
\end{figure}

\paragraph{Axis Refinement via Decomposition.}
\label{sec:refinement}
In preliminary analyses, we observed that although the five directions $\{\tilde{\mathbf{v}}^{\ell}_{k}\}_{k=1}^{5}$ capture distinct linguistic variations across dimensions, they appear to share a global offset trend, i.e., all axes tend to move activations from the ``neutral” region toward the  ``author-written” region, rather than changing in completely independent directions.
To verify this intuition, we take the paired neutral and original passages used for axis construction, average their last-token activations across layers and project them into a 3D PCA space.
As shown in Figure~\ref{fig:axis_analysis}(a), the neutral (blue) and author-written (red) samples form two compact clusters separated mainly along a single direction, revealing a strong global neutral$\rightarrow$author-written shift shared across dimensions.
Motivated by this finding, we propose to explicitly remove this shared component on a per-layer basis, in order to isolate axis-specific variations and improve the stability of multi-axis composition.

Concretely, for layer $\ell$, we stack the five axis vectors into
\begin{equation*}
\tilde{\mathbf{V}}^{\ell}
=\big[\tilde{\mathbf{v}}^{\ell}_{1},\tilde{\mathbf{v}}^{\ell}_{2},\cdots,\tilde{\mathbf{v}}^{\ell}_{5}\big]\in\mathbb{R}^{d\times 5},
\end{equation*}
and perform Singular Value Decomposition (SVD) $\tilde{\mathbf{V}}^{\ell}=\mathbf{U}^{\ell}\mathbf{\Sigma}^{\ell}{\mathbf{Q}^{\ell}}^{\top}$.
The first left singular vector $\mathbf{v}_O^{\ell}=\mathbf{U}^{\ell}_{:\!,1}\in\mathbb{R}^{d}$ corresponds to the most dominant direction of variation shared across all dimensions.
We identify this vector as the \textit{overall expressiveness direction}, which captures the collective tendency of activations to drift toward more stylized representations.
Simultaneously, we also compute its average magnitude $\rho^{\ell}_{O}$ by projecting the raw axes onto $\mathbf{v}_O^{\ell}$ to preserve the layer-wise intensity of this global trend.
Next, to emphasize the unique contribution of each axis, we remove this global trend from every $\tilde{\mathbf{v}}^{\ell}_{k}$, and obtain the \emph{refined unit axis} $\mathbf{v}^{\ell}_{k}$ with its magnitude $\rho^{\ell}_{k}$:

\begin{equation}
   \rho^{\ell}_{k} \cdot \mathbf{v}^{\ell}_{k} = \tilde{\mathbf{v}}^{\ell}_{k}-\mathbf{v}_O^{\ell}{\mathbf{v}_O^{\ell}}^{\!\top}\tilde{\mathbf{v}}^{\ell}_{k}.
\end{equation}
We retain the extracted magnitudes $\rho^{\ell}_{O}$ and $\boldsymbol{\rho}^{\ell} = [\rho^{\ell}_{1}, \dots, \rho^{\ell}_{5}]$ to restore the natural scale of interventions during the steering phase.
As shown in Figure~\ref{fig:axis_analysis}(b) and Appendix Figure~\ref{fig:pre_post_decomp_heatmap}, this decomposition step effectively reduces cross-axis correlations, and leads to more stable multi-axis combination.

\subsection{Authorial Coordinates Localization}
\label{localization}
After deriving the axes, we aim to localize a target book's position within the \textsc{LiteraryBigFive} space.
Let $\{x_{b,i}\}_{i=1}^{k}$ be $k$ reference passages from test book $b$. 
For layer $\ell$, let $a^{\ell}(x)\in\mathbb{R}^{d}$ denote the last-token activation and define its unit-normalized form $\widehat{a}^{\ell}(x)=\mathrm{norm}\!\big(a^{\ell}(x)\big)$.
We stack the refined axes into the basis matrix:
\[
\mathbf{V}^{\ell}=\big[\mathbf{v}^{\ell}_{1},\,\mathbf{v}^{\ell}_{2},\,\cdots,\,\mathbf{v}^{\ell}_{5}\big]\in\mathbb{R}^{d\times 5}.
\]
Since the inner product with unit axes provides a signed, scale-invariant measure of each dimension's intensity, for each reference passage, we can obtain per-layer authorial scores $\mathbf{s}^{\ell}_{b,i}$ by projecting its unit activation $\widehat{a}^{\ell}(x_{b,i})$ onto the five axes:
\begin{equation}
\mathbf{s}^{\ell}_{b,i}={\mathbf{V}^{\ell}}^\top\,\widehat{a}^{\ell}(x_{b,i})\in\mathbb{R}^{5}.   
\end{equation}

To interpret the style of the book and guide generation toward it, we aggregate these passage-level projections over $k$ references, yielding the book-level raw coordinates $\mathbf{s}_b^{\ell}$ at layer $\ell$:
\begin{equation}
 \mathbf{s}_b^{\ell}=\textstyle  \frac{1}{k}\sum_{i=1}^{k}\mathbf{s}^{\ell}_{b,i}\in\mathbb{R}^{5}.   
\end{equation}
These per-layer scores serve as personalized steering targets at the selected intervention layers~$\mathcal{L}$, capturing linguistic attributes ranging from local syntax to global semantics encoded at different depths~\cite{geva2021transformer}.
Furthermore, for analyzing the book's writing characteristics, we average the scores $\mathbf{s}_b^{\ell}$ across $\ell\in\mathcal{L}$ to derive book-level authorial coordinates:
\begin{equation}
 \mathbf{s}_b=\textstyle \frac{1}{|\mathcal{L}|}\sum_{\ell\in\mathcal{L}}\mathbf{s}_b^{\ell}\in\mathbb{R}^{5}.   
\end{equation}
Note that the coordinates above are not directly comparable across dimensions, as different axes exhibit varying dynamic ranges. 
To establish a consistent scale, we calibrate each dimension of $\mathbf{s}_b$ to a $[0,100]$ range using axis-specific anchor books (detailed procedure in Appendix~\ref{app:calibration}), enabling intuitive visualization of authorial profiles. 

\subsection{Interpretable Personalized Steering}
\label{steer}
After localizing the target book $b$’s authorial position in the \textsc{LiteraryBigFive} space as $\mathbf{s}_b$, we steer the model to rewrite an input passage $x$ into $\hat{x}$ so that its writing patterns align with the target authorial style while preserving the original semantic content.
Unlike prior methods that rely on a single direction per author~\cite{konen2024style,zhang2025personalized}, our method performs editing in an interpretable axis-aligned space with explicit and controllable per-dimension modulation.

Let $\mathbf{h}^{\ell}_{t}\!\in\!\mathbb{R}^{d}$ denote the hidden state at token $t$ and layer $\ell$, and let $\widehat{\mathbf{h}}^{\ell}_{t}=\mathrm{norm}\!\big(\mathbf{h}^{\ell}_{t}\big)$ be its normalized form. 
To stabilize steering, we augment the five-axis directions $\mathbf{V}^{\ell}$ with the shared expressiveness direction $\mathbf{v}^{\ell}_{O}$, which captures the global shift from neutral to literary style.
During generation, we project $\widehat{\mathbf{h}}^{\ell}_{t}$ onto both $\mathbf{V}^{\ell}$ and $\mathbf{v}^{\ell}_{O}$ to obtain the current layer-wise authorial scores:
\begin{equation*}
   \mathbf{s}^{\ell}_{t}\;=\;{\mathbf{V}^{\ell}}^{\!\top}\widehat{\mathbf{h}}^{\ell}_{t}\in\mathbb{R}^{5},
\quad
s^{\ell}_{O,t}\;=\;\big\langle \widehat{\mathbf{h}}^{\ell}_{t},\,\mathbf{v}^{\ell}_{O}\big\rangle . 
\end{equation*}
Similarly, the target overall expressiveness score is computed by averaging the projections of $k$ reference passages onto $\mathbf{v}^{\ell}_{O}$:
\begin{equation}
   s^{\ell}_{O,b}= \textstyle \frac{1}{k}\sum_{i=1}^{k}\big\langle \widehat{a}^{\ell}(x_{b,i}),\,\mathbf{v}^{\ell}_{O}\big\rangle. 
\end{equation}

Based on the current and target scores above, we can observe the \textit{\textbf{style gap}} $\mathbf{s}^{\ell}_{b}-\mathbf{s}^{\ell}_{t}$ (and $s^{\ell}_{O,b}-s^{\ell}_{O,t}$) between the current $t$-th token and the target author.
This gap indicates along which direction to move and by how much to bring the token closer to the target author in our \textsc{LiteraryBigFive} space.
We therefore convert it into edit strengths by rescaling it with the axis magnitudes $\boldsymbol{\rho}^{\ell}$ and $\rho^{\ell}_{O}$ extracted in the decomposition step:

\begin{equation}
\label{eq:alpha}
\begin{aligned}
\boldsymbol{\alpha}^{\ell}_{t}
&= \lambda\,\boldsymbol{\rho}^{\ell} \odot (\mathbf{s}^{\ell}_{b}-\mathbf{s}^{\ell}_{t}),\\
\alpha^{\ell}_{O,t}
&= \lambda\,\rho^{\ell}_{O}\,\big(s^{\ell}_{O,b}-s^{\ell}_{O,t}\big),
\end{aligned}
\end{equation}
where $\odot$ denotes element-wise product and $\lambda$ is a global control strength.
Finally, using these obtained coefficients, we steer the current token to the target author by updating its hidden state $\mathbf{h}^{\ell}_{t}\rightarrow {\mathbf{h}^{\ell}_{t}}^\prime$ for each selected intervention layer $\ell\in\mathcal{L}$:

\begin{equation}
{\mathbf{h}^{\ell}_{t}}^\prime
\;=\;
\mathbf{h}^{\ell}_{t}
\;+\;
\mathbf{V}^{\ell}\,\boldsymbol{\alpha}^{\ell}_{t}
\;+\;
\alpha^{\ell}_{O,t}\,\mathbf{v}^{\ell}_{O}.
\end{equation}

Edits proceed from shallow to deep layers, allowing style effects to accumulate across layers while avoiding over-correction at a single place.

\section{Experiments}

\begin{table*}[h]
\begin{small}
\begin{center}
\resizebox{\linewidth}{!}{
\begin{tabular}{l|ccccc|ccccc}
\toprule
\multirow{2}{*}{\textbf{Method}} &\multicolumn{5}{c|}{\textit{Reflections on the Revolution in France}} & \multicolumn{5}{c}{\textit{1984}} \\
 & ROUGE-1 & ROUGE-L & SIM & GPT-4 & Human & ROUGE-1 & ROUGE-L & SIM & GPT-4 & Human \\
\midrule
\textup{Few-shot}   & 38.0 & 28.7 & 73.9  & 64.6 & 42.5 & 55.1 & 46.6 & 87.1 & 69.7 & 54.5 \\
\textup{LLM-Steer} & 44.3 & 35.2 & 92.5 & 65.4 & 60.3 & 52.7 & 45.9 & 90.8 & 68.8 & 53.9 \\
\textup{LoRA} & 43.0 & 29.8 & 82.5 & 60.8 & 49.5 & 51.7 & 38.9 & 87.1 & 69.3 & 57.2 \\
\textup{ICV} & 43.6 & 33.8 & 93.6 & 65.4 & 60.3 & 54.7 & 48.6 & 92.8 & 73.1 & 71.5 \\
\textup{Mean-Centering} & 45.6 & 35.2 & 93.9 & 67.9 & 65.9 & 55.6 & 46.3 & 93.8 & 73.8 & 73.9 \\
\textup{CAA} & 41.6 & 31.9 & 89.7 & 59.7 & 45.8 & 43.3  & 35.9 & 85.7 & 57.5 & 46.7 \\
\textup{RepE} & 45.3 & 35.7  & 94.0 & 68.6 & 65.4 & 56.5 & 47.9 & 94.3 & 73.5 & 70.9 \\
\rowcolor{green!10}\textsc{LiteraryBigFive}& \textbf{46.1} & \textbf{36.4} & \textbf{94.4} & \textbf{69.4} & \textbf{69.2} &  \textbf{57.8} & \textbf{49.4} & \textbf{95.1} & \textbf{75.2} & \textbf{75.3} \\
\midrule
\multirow{2}{*}{\textbf{Method}} &\multicolumn{5}{c|}{\textit{Kidnapped}} & \multicolumn{5}{c}{\textit{Pride and Prejudice}} \\
 & ROUGE-1 & ROUGE-L & SIM & GPT-4 & Human & ROUGE-1 & ROUGE-L & SIM & GPT-4 & Human \\
\midrule
\textup{Few-shot} & 51.0 & 41.0  & 84.8  & 58.5 & 47.6 & 48.2 & 37.3 & 82.3 & 52.6 & 48.9 \\
\textup{LLM-Steer} & 50.3 & 43.3 & 91.2  & 58.8 & 57.1 & 44.8 & 37.0 & 90.3 & 55.2 & 51.5 \\
\textup{LoRA} & 53.3 & 44.9 & 92.2 & 58.7 & 56.4 & 49.1 & 34.1 & 87.5 & 56.1 & 53.7 \\
\textup{ICV} & 54.5 & 46.1 & 95.8 & 65.0 & 65.5 & 49.2 & 39.2 & 94.0 & 61.5 & 63.0 \\
\textup{Mean-Centering} & 55.7 & 47.0 & 96.1 & 66.5 & 61.8 & 50.9 & 40.0 & 94.6 & 64.0 & 66.7 \\
\textup{CAA} & 48.7 & 40.7 & 90.5 & 56.0 & 41.0 & 44.7 & 35.3 & 89.5 & 51.2 & 49.1 \\
\textup{RepE} & 55.9  & 47.6 & 96.4 & 65.9 & 68.3 & 50.2 & 40.2 & 94.4 & 63.5 & 67.8 \\
\rowcolor{green!10}\textsc{LiteraryBigFive} & \textbf{56.5} & \textbf{48.3} & \textbf{96.7}  & \textbf{67.8} & \textbf{73.8} & \textbf{51.3} & \textbf{41.7}  & \textbf{94.8} & \textbf{65.9} & \textbf{69.5} \\
\bottomrule
\end{tabular}
}
\end{center}
\end{small}
\vspace{-0.1in}
\caption{Experimental results on four books. For all metrics, higher scores indicate better performance. The best-performing methods are highlighted in \textbf{bold}, all results are scaled to 0-100 (two-tailed paired t-test, $p$\textless0.01).}
\vspace{-0.1in}
  \vspace{-3mm}
\label{tab:main_res}
\end{table*}

\subsection{Experimental Setup}

\textbf{Evaluation Data.} 
To evaluate \textsc{LiteraryBigFive} across diverse authors, we further curate a test set consisting of well-known books: Reflections on the Revolution in France by Edmund Burke, 1984 by George Orwell, Kidnapped by R. L. Stevenson, and Pride and Prejudice by Jane Austen. 
The resulting evaluation comprises 590 passage-level samples totaling 5,716 sentences, which substantially exceeds standard benchmarks such as the Shakespeare test set by \citet{xu2012paraphrasing}, containing 1.4 thousand sentences.
Each book possesses a distinct authorial expression, allowing for a comprehensive and rigorous assessment of our method's cross-author generalizability.

\textbf{Evaluation Metrics.}
Following previous works \cite{krishna2020reformulating,zhang2025personalized}, we adopt ROUGE-1/L \cite{lin-2004-rouge} to evaluate reconstruction quality by comparing generated passages against original texts of the target author.

To measure semantic preservation, we report embedding similarity (SIM), computed based on the cosine similarity of sentence representations encoded by the BGE model~\cite{chen2024m3}.

Beyond objective metrics, we leverage the strong capabilities of LLMs in evaluating complex writing characteristics~\cite{ostheimer2024text} by utilizing GPT-4 as a judge.
Specifically, we rate passages on a 0-10 scale considering two key dimensions: 
i) \textit{Authorial Adherence} measures how well the rewrite aligns with the target author's distinctive characteristics; and ii) \textit{Semantic Fidelity}, which evaluates the preservation of original meaning  (prompt in Appendix~\ref{app:eval_prompt}). 
To mitigate potential bias, we complement this with human evaluation, where two annotators rate responses using the same two-dimensional criteria.

\textbf{Baselines.}
We compare our \textsc{LiteraryBigFive} against various state-of-the-art baselines categorized as follows:
(1) Few-shot Prompting;
(2) Supervised Fine-Tuning, specifically LLM-Steer~\cite{han2024word}, which fine-tunes word embeddings via a linear transformation, and LoRA \cite{hu2022lora}, a parameter-efficient low-rank adaptation method;
(3) Activation Steering, including ICV~\cite{liu2024context}, Mean-Centering~\cite{jorgensen2023improving}, and CAA~\cite{rimsky2024steering}, RepE~\cite{zou2023transparency}.
Method introductions are detailed in Appendix~\ref{app:baseline}.


\textbf{Implementation Details.} 
We apply Llama2-7B-Chat~\cite{touvron2023llama2openfoundation} 
as the base LLM to implement our \textsc{LiteraryBigFive} and all baselines, with additional results on Qwen2.5-3B-Instruct \cite{qwen2.5} reported in Appendix~\ref{app:backbone_generalization}.
All experiments were conducted with NVIDIA RTX 5880 Ada GPUs. 
Detailed Hyperparameters setting are provided in the Appendix \ref{app:implementation}.

\subsection{Main Results}

As shown in Table \ref{tab:main_res}, across four stylistically diverse books, \textsc{LiteraryBigFive} consistently outperforms all baselines on ROUGE, SIM, GPT-4 and human evaluations. We summarize three key observations.
\textit{(1) Interpretable, axis-aligned editing yields the strongest and most stable personalized generation}.
Unlike conventional editing methods that operate in entangled latent spaces, \textsc{LiteraryBigFive} leverages interpretable BigFive axes to support context-aware and fine-grained author-specific adjustment (with qualitative examples provided in Appendix~\ref{app:case_study}), consistently achieving higher ROUGE scores while preserving semantic fidelity and stylistic coherence.
\textit{(2) Robustness across diverse authors.} The evaluation ranges
from Burke's political rhetoric to Austen's narrative prose. While baseline performance fluctuates, \textsc{LiteraryBigFive} maintains high scores across all domains. 
This indicates that our model generalizes well to different styles without overfitting to specific corpora.
\begin{figure}
\centering
\includegraphics[width=\columnwidth]{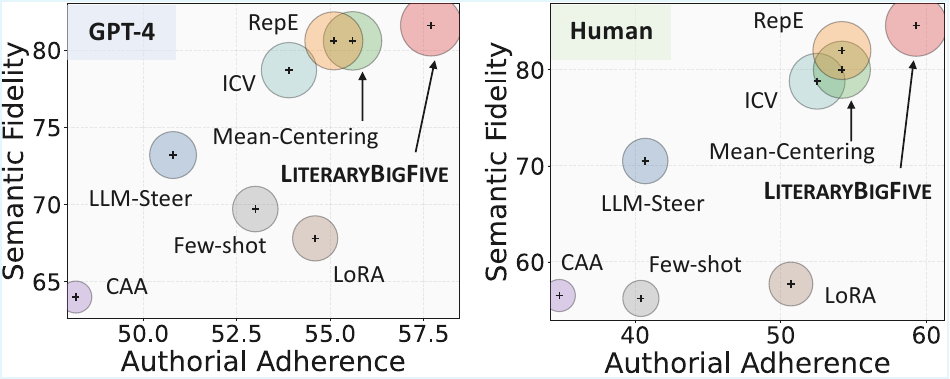}
  \caption{Performance analysis of Semantic Fidelity and Authorial Adherence. Radius denotes mean value.}
  \label{fig:sasf}
    \vspace{-1.5mm}
\end{figure}
\begin{table}[tbp]
\centering
\small
\resizebox{\linewidth}{!}{
\begin{tabular}{@{}l|cccc@{}}
\toprule
Variants & ROUGE-1 & ROUGE-L & SIM & GPT-4 \\ \midrule
-w/o Decomposition & 52.1 & 43.2 & 94.7 & 68.8   \\ 
-w/o Style Gap & 49.3 & 40.0 & 91.6 &  66.6  \\ 
\textsc{LiteraryBigFive} & \textbf{53.0} & \textbf{44.0} & \textbf{95.2} &  \textbf{69.6}  \\ 
\bottomrule
\end{tabular}}
\caption{Ablation study on \textsc{LiteraryBigFive}.
\textbf{Bold} numbers indicate statistically significant improvements over ablations
(two-tailed paired t-test, $p$\textless0.01). }
\label{tab:ablation}
  \vspace{-1.5mm}
\end{table}
\textit{(3) High authorial adherence with robust semantic preservation.}
A major challenge in personalized generation is aligning the output with a target author's writing characteristics without altering the original meaning.
As visualized in Figure~\ref{fig:sasf}, \textsc{LiteraryBigFive} occupies the optimal region (top-right), achieving the highest scores on both dimensions simultaneously.
This demonstrates that our approach effectively disentangles authorial expression from content, enabling faithful personalization while preserving core semantics.
Human annotators achieve a Cohen’s $\kappa$ of 0.59, demonstrating moderate inter-annotator agreement.
It can also be observed that GPT-4’s scores closely align with human evaluations, supporting the reliability of LLM-based assessment.

\section{Analysis and Discussion}

\begin{table*}[h]
\centering
\small
\renewcommand{\arraystretch}{1.05}
\resizebox{\textwidth}{!}{%
    \begin{tabular}{l c m{12.7cm}} 
    \toprule
    \textbf{Dimension} & \textbf{Strength} & \textbf{Generated Text Snippet } \\
    \midrule
    \multirow{2}{*}{\textbf{Classicism}} 
    & -0.8 & ...persons... who had \styleLow{caused resentment} towards the throne by \styleLow{accepting its generous rewards}... \\
    & 0.8 & ...persons... who had \styleHigh{brought an odium} on the throne by the \styleHigh{prodigal dispensation of its bounties}... \\
    \rowcolor{gray!10} 
    \multicolumn{3}{p{16.5cm}}{ 
        \raggedright 
        \textit{\textbf{Analysis:}} High Classicism steers towards \textit{Archaic Lexicon}. Note the shift from modern ``caused resentment'' to Latinate \styleHigh{odium}, and from simple ``generous rewards'' to more period-specific phrasing \styleHigh{prodigal dispensation}.
    } \\
    \midrule
    \multirow{2}{*}{\textbf{Emotionality}} 
    & -0.8 & ...Kitty was not completely surprised. \styleLow{I am very sorry. It is an imprudent match} for both of them! But I hope for the best... \\
    & 0.8 & ...Kitty... does not seem so wholly unexpected. \styleHigh{Our poor mother is sadly grieved. So imprudent a match} on both sides! But I am willing to hope... \\
    \rowcolor{gray!10}
    \multicolumn{3}{p{16.5cm}}{
        \raggedright
        \textit{\textbf{Analysis:}} High Emotionality drives \textit{Affective Intensity}. The text shifts from neutral observation to personal sentiment, adding emotional weight through words like \styleHigh{sadly grieved} and emphatic structures (``So imprudent...'').
    } \\
    \midrule
    \multirow{2}{*}{\textbf{Ornateness}} 
    & -0.8 & ...She is \styleLow{friendly and gracious}, and she will probably \styleLow{pay some attention} to you... \\
    & 0.8 & ...She is \styleHigh{all affability and condescension}, and I doubt not but you will be \styleHigh{honoured with some portion of her notice}... \\
    \rowcolor{gray!10}
    \multicolumn{3}{p{16.5cm}}{
        \raggedright
        \textit{\textbf{Analysis:}} High Ornateness promotes \textit{Syntactic Complexity}. Straightforward adjectives like (``friendly'') are elaborated into abstract noun phrases (\styleHigh{affability and condescension}), resulting in a more decorative and indirect writing style.
    } \\
    \bottomrule
    \end{tabular}%
}
    \caption{Qualitative comparison of observed shifts, with linguistic analysis highlighted in shaded rows. 
We present cases with steering strengths $\alpha \in \{-0.8, +0.8\}$ here, while more results and analysis are available in Appendix~\ref{app:style_parallel_analysis}.}
\label{tab:style_parallel_analysis}
\vspace{-2mm}
\end{table*}

\subsection{Ablation Study}

We also conduct an ablation study to examine the contribution of each key component in \textsc{LiteraryBigFive}, as shown in Table~\ref{tab:ablation}. 
First, removing the axis decomposition step in \S\ref{sec:refinement} and using raw book-level directions (\textit{-w/o Decomposition}) leads to a noticeable drop across all metrics, indicating that refinement is essential for isolating clean, dimension-specific authorial signals. 
Furthermore, disabling the dynamic adaption of the style gap in \S\ref{localization} and applying a fixed steering strength (\textit{-w/o Style Gap}) yields an even larger performance degradation than removing refinement.
This highlights the crucial role of adaptive token-level steering, as authorial cues are unevenly distributed across a passage and require context-sensitive adjustment to avoid insufficient or excessive intervention.
Overall, these findings validate that combining axis refinement with adaptive steering is necessary to achieve optimal personalization and semantic preservation.


\subsection{Authorial Coordinates Analysis}

To assess the interpretability of authorial coordinates derived in the \textsc{LiteraryBigFive} space, we evaluate their alignment with independent stylistic judgements produced by frontier LLMs.

\begin{figure}[t]
\centering
  \includegraphics[width=\columnwidth]{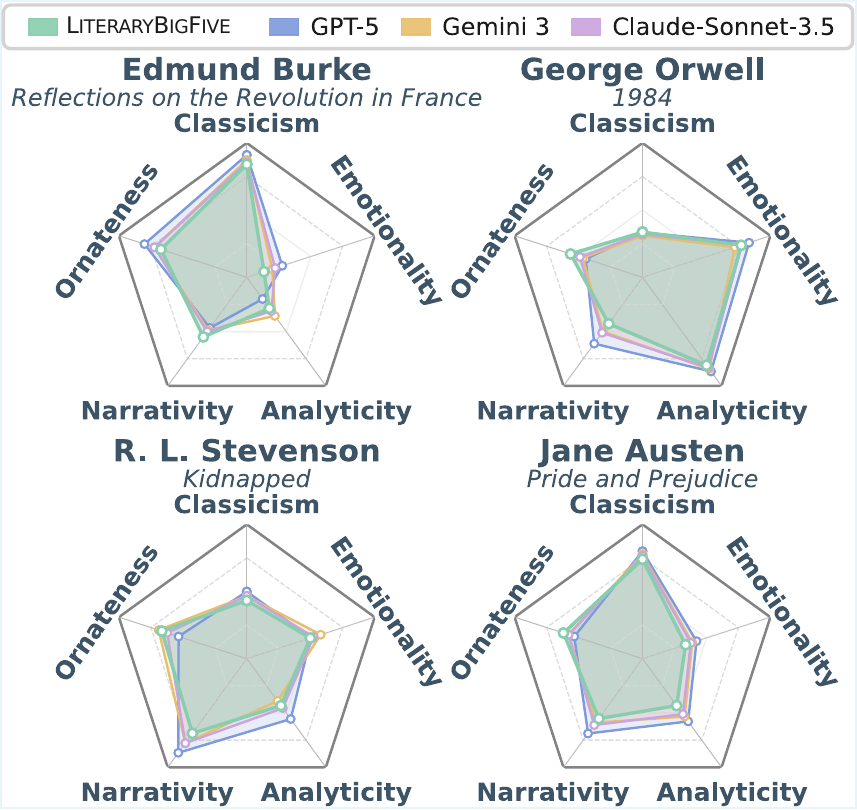}
  \caption{Radar charts comparing \textsc{LiteraryBigFive} coordinates with LLM-based authorial scores across four authors. The strong overlap indicates consistent authorial stylistic characterization.}
  \label{fig:radar}
\end{figure}

Accordingly, we ask GPT-5, Claude~3.5, and Gemini~3 to rate each book on a 0--100 scale along the five dimensions defined in \textsc{LiteraryBigFive} (prompt in Appendix~\ref{app:assessment}).
We then compute the Pearson correlation between our model coordinates and the ensemble average of the LLM scores.
Results in Appendix~\ref{app:style_score} show strong alignment between \textsc{LiteraryBigFive} and the LLM consensus, with an average Pearson correlation of $r=0.96$ across all axes.
As shown in the radar charts (Figure~\ref{fig:radar}), our method captures stylistic patterns consistent with advanced LLM judgments.
For example, the high \textit{Classicism} of Edmund Burke and the high \textit{Analyticity} of George Orwell are reflected in both our coordinates and the LLM ratings.
Discrepancies in the radar plots further improve interpretability by revealing dimensions where our model diverges from the LLM consensus.

\vspace{-1mm}
\subsection{Case Study: Dimension Steering}


To explore the interpretability of \textsc{LiteraryBigFive} and gain qualitative insight into individual dimensions, we conduct a case study examining stylistic shifts induced by steering along a single dimension.
Specifically, we randomly sample 40 texts from our test set.
For each text, we apply the axis to one target dimension at a time while keeping other dimensions at zero to ensure isolation, and generate rewrites by varying the steering strength $\alpha \in \{-0.8, -0.4, 0, +0.4, +0.8\}$. 
As reflected in Table~\ref{tab:style_parallel_analysis} (full version is in Appendix~\ref{app:style_parallel_analysis}), the results show that BigFive axes effectively modulate corresponding dimensions, such as the transition to Latinate diction in Classicism, without changing the underlying meaning of the text.

\section{Conclusion}

We present \textsc{LiteraryBigFive}, a framework that reframes isolated author's writing characteristics into a unified and interpretable five-dimensional space.
By leveraging a \textit{localize-and-steer} mechanism, our approach integrates precise, interpretable analysis of authorial expression with low-cost personalized generation for new authors.
Experimental results demonstrate that \textsc{LiteraryBigFive} outperforms baselines in authorial expressiveness and semantic fidelity, while the derived coordinates closely match established literary consensus.
Future work may extend this paradigm to multilingual settings and interactive writing support systems.

\section*{Limitations}

Despite the effectiveness of our framework, we acknowledge specific constraints in its design and application.
First, the axes in this study are primarily derived from English literary classics, which reflects the currently limited exploration of this task within the broader research landscape. We expect future work to extend this approach to richer linguistic and literary settings.
Second, the steering mechanism operates globally on the residual stream layers. 
While this approach effectively captures holistic writing attributes, it lacks the granularity required to manipulate specific long-range dependencies, which might be better addressed by targeting individual attention heads or specific components.
Finally, our method relies on the extraction and manipulation of internal activation vectors. This dependency on white-box access limits the framework's applicability to open-weight models and prevents it from working with closed-source language model APIs that do not provide direct access to these internal embeddings.

\section*{Ethical Considerations}
We prioritize the responsible development of personalized text generation frameworks and strictly adhere to ethical guidelines regarding data usage and model deployment. 
All datasets used in our experiments are derived from publicly available sources, primarily consisting of literary works in the public domain, and no private, sensitive, or personally identifiable information is included. 
Consequently, the generation process follows open and reproducible settings without targeting any real individuals. 
While our framework enables the adaptation of writing characteristics, we acknowledge the potential risks associated with automated author imitation, such as non-consensual impersonation or the generation of misleading content. 
We emphasize that \textsc{LiteraryBigFive} is designed specifically for creative support, literary analysis, and adaptive assistance; by grounding our method in interpretable literary dimensions, we aim to foster transparency in how text style is manipulated. 

\bibliography{custom_simplified}

\begin{thebibliography}{62}
\providecommand{\natexlab}[1]{#1}

\bibitem[{Abbott(2021)}]{abbott2021narrative}
H~Porter Abbott. 2021.
\newblock \emph{The Cambridge introduction to narrative}.
\newblock Cambridge University Press.

\bibitem[{Auerbach(2013)}]{auerbach2013mimesis}
Erich Auerbach. 2013.
\newblock \emph{Mimesis: The Representation of Reality in Western Literature}.
\newblock Princeton University Press.

\bibitem[{Banayeeanzade et~al.(2026)Banayeeanzade, Tak, Bahrani, Bolourani, Blas, Ferrara, Gratch, and Karimireddy}]{abs-2510-04484}
Amin Banayeeanzade, Ala~N. Tak, Fatemeh Bahrani, Anahita Bolourani, Leonardo Blas, Emilio Ferrara, Jonathan Gratch, and Sai~Praneeth Karimireddy. 2026.
\newblock Psychological steering in {LLM}s: An evaluation of effectiveness and trustworthiness.
\newblock In \emph{Proceedings of ACL}.

\bibitem[{Bhandarkar et~al.(2024)Bhandarkar, Wilson, Swarup, and Woodard}]{bhandarkar2024emulating}
Avanti Bhandarkar, Ronald Wilson, Anushka Swarup, and Damon Woodard. 2024.
\newblock Emulating author style: A feasibility study of prompt-enabled text stylization with off-the-shelf {LLM}s.
\newblock In \emph{Proceedings of the 1st Workshop on Personalization of Generative AI Systems (PERSONALIZE 2024)}.

\bibitem[{Biber(1991)}]{biber1991variation}
Douglas Biber. 1991.
\newblock \emph{Variation across speech and writing}.
\newblock Cambridge university press.

\bibitem[{Biber(1995)}]{biber1995dimensions}
Douglas Biber. 1995.
\newblock \emph{Dimensions of register variation: A cross-linguistic comparison}.
\newblock Cambridge University Press.

\bibitem[{Biber and Conrad(2019)}]{BiberConrad2019Register}
Douglas Biber and Susan Conrad. 2019.
\newblock \emph{Register, Genre, and Style}.
\newblock Cambridge University Press, New York.

\bibitem[{Biber and Gray(2016)}]{biber2016grammaticalcomplexity}
Douglas Biber and Bethany Gray. 2016.
\newblock \emph{Grammatical complexity in academic English: Linguistic change in writing}.
\newblock Cambridge University Press.

\bibitem[{Booth(1983)}]{booth1983rhetoric}
Wayne~C Booth. 1983.
\newblock \emph{The rhetoric of fiction}.
\newblock University of Chicago Press.

\bibitem[{Boyd et~al.(2022)Boyd, Ashokkumar, Seraj, and Pennebaker}]{boyd2022development}
Ryan~L Boyd, Ashwini Ashokkumar, Sarah Seraj, and James~W Pennebaker. 2022.
\newblock The development and psychometric properties of liwc-22.
\newblock \emph{Austin, TX: University of Texas at Austin}.

\bibitem[{Chen et~al.(2024)Chen, Xiao, Zhang, Luo, Lian, and Liu}]{chen2024m3}
Jianlyu Chen, Shitao Xiao, Peitian Zhang, Kun Luo, Defu Lian, and Zheng Liu. 2024.
\newblock {M}3-embedding: Multi-linguality, multi-functionality, multi-granularity text embeddings through self-knowledge distillation.
\newblock In \emph{Proceedings of ACL Findings}.

\bibitem[{Chen et~al.(2025)Chen, Arditi, Sleight, Evans, and Lindsey}]{abs-2507-21509}
Runjin Chen, Andy Arditi, Henry Sleight, Owain Evans, and Jack Lindsey. 2025.
\newblock \href {https://arxiv.org/abs/2507.21509} {Persona vectors: Monitoring and controlling character traits in language models}.
\newblock \emph{Preprint}, arXiv:2507.21509.

\bibitem[{Cheng et~al.(2025)Cheng, Gan, Jiang, Wang, Yin, Luo, Fu, and Gu}]{abs-2508-17621}
Zifeng Cheng, Jinwei Gan, Zhiwei Jiang, Cong Wang, Yafeng Yin, Xiang Luo, Yuchen Fu, and Qing Gu. 2025.
\newblock Steering when necessary: Flexible steering large language models with backtracking.
\newblock In \emph{Proceedings of NeurIPS}.

\bibitem[{Eliot(2024)}]{eliot2024sacred}
Thomas~Stearns Eliot. 2024.
\newblock \emph{The Sacred Wood, Essays on Poetry and Criticism}.
\newblock Otbebookpublishing.

\bibitem[{Faulkner(1956)}]{faulkner1956william}
William Faulkner. 1956.
\newblock William faulkner, the art of fiction no. 12.
\newblock \emph{The Paris Review}.

\bibitem[{Gao et~al.(2026)Gao, Zhang, Liu, Ji, Wang, Song, Ghosh, Mohamed, Nakov, and Chen}]{gao2026the}
Lang Gao, Jinghui Zhang, Wei Liu, Fengxian Ji, Chenxi Wang, Zirui Song, Akash Ghosh, Youssef Mohamed, Preslav Nakov, and Xiuying Chen. 2026.
\newblock The cylindrical representation hypothesis for language model steering.
\newblock In \emph{Proceedings of ICML}.

\bibitem[{Geva et~al.(2021)Geva, Schuster, Berant, and Levy}]{geva2021transformer}
Mor Geva, Roei Schuster, Jonathan Berant, and Omer Levy. 2021.
\newblock Transformer feed-forward layers are key-value memories.
\newblock In \emph{Proceedings of EMNLP}.

\bibitem[{Goldberg(1993)}]{goldberg1993structure}
Lewis~R Goldberg. 1993.
\newblock The structure of phenotypic personality traits.
\newblock \emph{American psychologist}.

\bibitem[{Han et~al.(2024)Han, Xu, Li, Fung, Sun, Jiang, Abdelzaher, and Ji}]{han2024word}
Chi Han, Jialiang Xu, Manling Li, Yi~Fung, Chenkai Sun, Nan Jiang, Tarek Abdelzaher, and Heng Ji. 2024.
\newblock Word embeddings are steers for language models.
\newblock In \emph{Proceedings of ACL}.

\bibitem[{Hemingway(1999)}]{hemingway1999death}
Ernest Hemingway. 1999.
\newblock \emph{Death in the Afternoon}.
\newblock Simon and Schuster.

\bibitem[{Henkle(1970)}]{henkle1970boundaries}
Roger Henkle. 1970.
\newblock The boundaries of fiction: Carlyle, macaulay, newman.
\newblock In \emph{Novel: A Forum on Fiction}.

\bibitem[{Holmes(1998)}]{holmes1998evolution}
David~I Holmes. 1998.
\newblock The evolution of stylometry in humanities scholarship.
\newblock \emph{Literary and linguistic computing}.

\bibitem[{Hu et~al.(2022)Hu, Shen, Wallis, Allen-Zhu, Li, Wang, Wang, and Chen}]{hu2022lora}
Edward~J Hu, Yelong Shen, Phillip Wallis, Zeyuan Allen-Zhu, Yuanzhi Li, Shean Wang, Lu~Wang, and Weizhu Chen. 2022.
\newblock Lo{RA}: Low-rank adaptation of large language models.
\newblock In \emph{Proceedings of ICLR}.

\bibitem[{Hu et~al.(2017)Hu, Yang, Liang, Salakhutdinov, and Xing}]{hu2017toward}
Zhiting Hu, Zichao Yang, Xiaodan Liang, Ruslan Salakhutdinov, and Eric~P. Xing. 2017.
\newblock Toward controlled generation of text.
\newblock In \emph{Proceedings of ICML}.

\bibitem[{Ishiguro(2008)}]{ishiguro2007art}
Kazuo Ishiguro. 2008.
\newblock The art of fiction no. 196.
\newblock \emph{Paris Review}.

\bibitem[{Jhamtani et~al.(2017)Jhamtani, Gangal, Hovy, and Nyberg}]{jhamtani2017shakespearizing}
Harsh Jhamtani, Varun Gangal, Eduard Hovy, and Eric Nyberg. 2017.
\newblock Shakespearizing modern language using copy-enriched sequence to sequence models.
\newblock In \emph{Proceedings of the Workshop on Stylistic Variation}.

\bibitem[{Jiang et~al.(2024)Jiang, Zhang, Cao, Breazeal, Roy, and Kabbara}]{jiang2024personallm}
Hang Jiang, Xiajie Zhang, Xubo Cao, Cynthia Breazeal, Deb Roy, and Jad Kabbara. 2024.
\newblock {P}ersona{LLM}: Investigating the ability of large language models to express personality traits.
\newblock In \emph{Proceedings of NAACL Findings}.

\bibitem[{John and Srivastava(1999)}]{john1999bigfive}
Oliver~P. John and Sanjay Srivastava. 1999.
\newblock The big five trait taxonomy: History, measurement, and theoretical perspectives.
\newblock In Lawrence~A. Pervin and Oliver~P. John, editors, \emph{Handbook of Personality: Theory and Research}, 2 edition, pages 102--138. Guilford Press, New York.

\bibitem[{Jorgensen et~al.(2023)Jorgensen, Cope, Schoots, and Shanahan}]{jorgensen2023improving}
Ole Jorgensen, Dylan Cope, Nandi Schoots, and Murray Shanahan. 2023.
\newblock \href {https://arxiv.org/abs/2312.03813} {Improving activation steering in language models with mean-centring}.
\newblock \emph{Preprint}, arXiv:2312.03813.

\bibitem[{Kim et~al.(2018)Kim, Wattenberg, Gilmer, Cai, Wexler, Vi{\'{e}}gas, and Sayres}]{KimWGCWVS18}
Been Kim, Martin Wattenberg, Justin Gilmer, Carrie~J. Cai, James Wexler, Fernanda~B. Vi{\'{e}}gas, and Rory Sayres. 2018.
\newblock Interpretability beyond feature attribution: Quantitative testing with concept activation vectors {(TCAV)}.
\newblock In \emph{Proceedings of ICML}.

\bibitem[{Konen et~al.(2024)Konen, Jentzsch, Diallo, Sch{\"u}tt, Bensch, El~Baff, Opitz, and Hecking}]{konen2024style}
Kai Konen, Sophie Jentzsch, Diaoul{\'e} Diallo, Peer Sch{\"u}tt, Oliver Bensch, Roxanne El~Baff, Dominik Opitz, and Tobias Hecking. 2024.
\newblock Style vectors for steering generative large language models.
\newblock In \emph{Proceedings of EACL Findings}.

\bibitem[{Krishna et~al.(2020)Krishna, Wieting, and Iyyer}]{krishna2020reformulating}
Kalpesh Krishna, John Wieting, and Mohit Iyyer. 2020.
\newblock Reformulating unsupervised style transfer as paraphrase generation.
\newblock In \emph{Proceedings of EMNLP}.

\bibitem[{Kuiken and Jacobs(2021)}]{kuiken2021handbook}
Donald Kuiken and Arthur~M. Jacobs, editors. 2021.
\newblock \emph{Handbook of Empirical Literary Studies}.
\newblock De Gruyter, Berlin and Boston.

\bibitem[{Labov(1972)}]{labov1972language}
William Labov. 1972.
\newblock \emph{Language in the inner city: Studies in the Black English vernacular}.
\newblock University of Pennsylvania Press.

\bibitem[{Lanham(1991)}]{lanham1991handlist}
Richard~A Lanham. 1991.
\newblock \emph{A handlist of rhetorical terms}.
\newblock University of California Press.

\bibitem[{Levine(1968)}]{levine1968carlylese}
George Levine. 1968.
\newblock The use and abuse of carlylese.
\newblock In George Levine and William~A. Madden, editors, \emph{The Art of Victorian Prose}, pages 101--126. Oxford University Press, New York.

\bibitem[{Lin(2004)}]{lin-2004-rouge}
Chin-Yew Lin. 2004.
\newblock {ROUGE}: A package for automatic evaluation of summaries.
\newblock In \emph{Text Summarization Branches Out}.

\bibitem[{Liu et~al.(2024)Liu, Ye, Xing, and Zou}]{liu2024context}
Sheng Liu, Haotian Ye, Lei Xing, and James~Y Zou. 2024.
\newblock In-context vectors: Making in context learning more effective and controllable through latent space steering.
\newblock In \emph{Proceedings of ICML}.

\bibitem[{Ma et~al.(2025)Ma, Xu, Lin, Wang, Chu, Gao, Zhao, and Wang}]{ma2025dressing}
Xinyu Ma, Yifeng Xu, Yang Lin, Tianlong Wang, Xu~Chu, Xin Gao, Junfeng Zhao, and Yasha Wang. 2025.
\newblock {DRESS}ing up {LLM}: Efficient stylized question-answering via style subspace editing.
\newblock In \emph{Proceedings of ICLR}.

\bibitem[{Martin and White(2005)}]{martin2003language}
James~R. Martin and Peter R.~R. White. 2005.
\newblock \emph{The Language of Evaluation: Appraisal in English}.
\newblock Palgrave Macmillan.

\bibitem[{McKeon(2002)}]{mckeon2002origins}
Michael McKeon. 2002.
\newblock \emph{The origins of the English novel, 1600-1740}.
\newblock JHU Press.

\bibitem[{Ning et~al.(2025)Ning, Liu, Wu, Wu, Berlowitz, Prakash, Green, O'Banion, and Xie}]{lin2025userllm}
Lin Ning, Luyang Liu, Jiaxing Wu, Neo Wu, Devora Berlowitz, Sushant Prakash, Bradley Green, Shawn O'Banion, and Jun Xie. 2025.
\newblock User-llm: Efficient llm contextualization with user embeddings.
\newblock In \emph{Companion Proceedings of the ACM on Web Conference 2025}.

\bibitem[{Niven(1978)}]{niven1978dh}
Alistair Niven. 1978.
\newblock \emph{D.H. Lawrence: the novels}.
\newblock Cambridge University Press.

\bibitem[{OpenAI et~al.(2024)OpenAI, Achiam, Adler, and et~al.}]{openai2024gpt4technicalreport}
OpenAI, Josh Achiam, Steven Adler, and et~al. 2024.
\newblock \href {https://arxiv.org/abs/2303.08774} {{GPT}-4 technical report}.
\newblock \emph{Preprint}, arXiv:2303.08774.

\bibitem[{Ostheimer et~al.(2024)Ostheimer, Nagda, Kloft, and Fellenz}]{ostheimer2024text}
Phil Ostheimer, Mayank Nagda, Marius Kloft, and Sophie Fellenz. 2024.
\newblock Text style transfer evaluation using large language models.
\newblock In \emph{Proceedings of LREC-COLING}.

\bibitem[{Pater(2023)}]{pater2023renaissance}
Walter Pater. 2023.
\newblock \emph{The Renaissance: studies in art and poetry}.
\newblock Univ of California Press.

\bibitem[{Prabhumoye et~al.(2018)Prabhumoye, Tsvetkov, Salakhutdinov, and Black}]{prabhumoye2018style}
Shrimai Prabhumoye, Yulia Tsvetkov, Ruslan Salakhutdinov, and Alan~W Black. 2018.
\newblock Style transfer through back-translation.
\newblock In \emph{Proceedings of ACL}.

\bibitem[{Qin et~al.(2025)Qin, Zhu, Fan, and Hui}]{qin2025writingsupport}
Hua~Xuan Qin, Guangzhi Zhu, Mingming Fan, and Pan Hui. 2025.
\newblock Toward personalizable ai node graph creative writing support: Insights on preferences for generative ai features and information presentation across story writing processes.
\newblock In \emph{Proceedings of the 2025 CHI Conference on Human Factors in Computing Systems}.

\bibitem[{Reif et~al.(2022)Reif, Ippolito, Yuan, Coenen, Callison-Burch, and Wei}]{reif2022recipe}
Emily Reif, Daphne Ippolito, Ann Yuan, Andy Coenen, Chris Callison-Burch, and Jason Wei. 2022.
\newblock A recipe for arbitrary text style transfer with large language models.
\newblock In \emph{Proceedings of ACL}.

\bibitem[{Rimsky et~al.(2024)Rimsky, Gabrieli, Schulz, Tong, Hubinger, and Turner}]{rimsky2024steering}
Nina Rimsky, Nick Gabrieli, Julian Schulz, Meg Tong, Evan Hubinger, and Alexander Turner. 2024.
\newblock Steering {Llama} 2 via contrastive activation addition.
\newblock In \emph{Proceedings of ACL}.

\bibitem[{Touvron et~al.(2023)Touvron, Martin, Stone, and et~al.}]{touvron2023llama2openfoundation}
Hugo Touvron, Louis Martin, Kevin Stone, and et~al. 2023.
\newblock \href {https://arxiv.org/abs/2307.09288} {Llama 2: Open foundation and fine-tuned chat models}.
\newblock \emph{Preprint}, arXiv:2307.09288.

\bibitem[{Vickers(1968)}]{vickers1968francis}
Brian Vickers. 1968.
\newblock \emph{Francis Bacon and renaissance prose}.
\newblock Cambridge University Press.

\bibitem[{Wang et~al.(2024)Wang, Peng, Que, Liu, Zhou, Wu, Guo, Gan, Ni, Yang, Zhang, Zhang, Ouyang, Xu, Huang, Fu, and Peng}]{wang2024rolellm}
Noah Wang, Z.Y. Peng, Haoran Que, Jiaheng Liu, Wangchunshu Zhou, Yuhan Wu, Hongcheng Guo, Ruitong Gan, Zehao Ni, Jian Yang, Man Zhang, Zhaoxiang Zhang, Wanli Ouyang, Ke~Xu, Wenhao Huang, Jie Fu, and Junran Peng. 2024.
\newblock {R}ole{LLM}: Benchmarking, eliciting, and enhancing role-playing abilities of large language models.
\newblock In \emph{Proceedings of ACL Findings}.

\bibitem[{Watt(1957)}]{watt1957rise}
Ian Watt. 1957.
\newblock \emph{The Rise of the Novel: Studies in Defoe, Richardson and Fielding}.
\newblock University of California Press, Berkeley.

\bibitem[{Wimsatt(1941)}]{wimsatt1941prose}
W.K. Wimsatt. 1941.
\newblock \emph{The Prose Style of Samuel Johnson}.
\newblock Yale University Press.

\bibitem[{Xu et~al.(2012)Xu, Ritter, Dolan, Grishman, and Cherry}]{xu2012paraphrasing}
Wei Xu, Alan Ritter, Bill Dolan, Ralph Grishman, and Colin Cherry. 2012.
\newblock Paraphrasing for style.
\newblock In \emph{Proceedings of COLING}.

\bibitem[{Yang et~al.(2025)Yang, Yang, Zhang, Hui, Zheng, Yu, Li, Liu, Huang, Wei, Lin, Yang, Tu, Zhang, Yang, Yang, Zhou, Lin, Dang, Lu, Bao, Yang, Yu, Li, Xue, Zhang, Zhu, Men, Lin, Li, Tang, Xia, Ren, Ren, Fan, Su, Zhang, Wan, Liu, Cui, Zhang, and Qiu}]{qwen2.5}
An~Yang, Baosong Yang, Beichen Zhang, Binyuan Hui, Bo~Zheng, Bowen Yu, Chengyuan Li, Dayiheng Liu, Fei Huang, Haoran Wei, Huan Lin, Jian Yang, Jianhong Tu, Jianwei Zhang, Jianxin Yang, Jiaxi Yang, Jingren Zhou, Junyang Lin, Kai Dang, and 23 others. 2025.
\newblock \href {https://arxiv.org/abs/2412.15115} {Qwen2.5 technical report}.
\newblock \emph{Preprint}, arXiv:2412.15115.

\bibitem[{Yu et~al.(2024)Yu, Zang, Wang, Zhuang, and Gu}]{yu2024charpoet}
Chengyue Yu, Lei Zang, Jiaotuan Wang, Chenyi Zhuang, and Jinjie Gu. 2024.
\newblock {C}har{P}oet: A {C}hinese classical poetry generation system based on token-free {LLM}.
\newblock In \emph{Proceedings of ACL}.

\bibitem[{Zhang et~al.(2025{\natexlab{a}})Zhang, Liu, Wang, Liu, Wu, Wang, and Chua}]{zhang2025personalized}
Jinghao Zhang, Yuting Liu, Wenjie Wang, Qiang Liu, Shu Wu, Liang Wang, and Tat-Seng Chua. 2025{\natexlab{a}}.
\newblock Personalized text generation with contrastive activation steering.
\newblock In \emph{Proceedings of ACL}.

\bibitem[{Zhang et~al.(2025{\natexlab{b}})Zhang, Wan, Xu, Li, Liu, and Chen}]{zhang2025fromindividual}
Jinghui Zhang, Kaiyang Wan, Longwei Xu, Ao~Li, Zongfang Liu, and Xiuying Chen. 2025{\natexlab{b}}.
\newblock From individuals to crowds: Dual-level public response prediction in social media.
\newblock In \emph{Proceedings of ACM MM}.

\bibitem[{Zhang et~al.(2025{\natexlab{c}})Zhang, Rossi, Kveton, Shao, Yang, Zamani, Dernoncourt, Barrow, Yu, Kim, Zhang, Gu, Derr, Chen, Wu, Chen, Wang, Mitra, Lipka, Ahmed, and Wang}]{zhang2025personalization}
Zhehao Zhang, Ryan~A. Rossi, Branislav Kveton, Yijia Shao, Diyi Yang, Hamed Zamani, Franck Dernoncourt, Joe Barrow, Tong Yu, Sungchul Kim, Ruiyi Zhang, Jiuxiang Gu, Tyler Derr, Hongjie Chen, Junda Wu, Xiang Chen, Zichao Wang, Subrata Mitra, Nedim Lipka, and 2 others. 2025{\natexlab{c}}.
\newblock Personalization of large language models: A survey.
\newblock \emph{TMLR}.

\bibitem[{Zou et~al.(2023)Zou, Phan, Chen, Campbell, Guo, Ren, Pan, Yin, Mazeika, Dombrowski, Goel, Li, Byun, Wang, Mallen, Basart, Koyejo, Song, Fredrikson, Kolter, and Hendrycks}]{zou2023transparency}
Andy Zou, Long Phan, Sarah Chen, James Campbell, Phillip Guo, Richard Ren, Alexander Pan, Xuwang Yin, Mantas Mazeika, Ann-Kathrin Dombrowski, Shashwat Goel, Nathaniel Li, Michael~J. Byun, Zifan Wang, Alex Mallen, Steven Basart, Sanmi Koyejo, Dawn Song, Matt Fredrikson, and 2 others. 2023.
\newblock \href {https://arxiv.org/abs/2310.01405} {Representation engineering: A top-down approach to ai transparency}.
\newblock \emph{Preprint}, arXiv:2310.01405.

\end{thebibliography}

\newpage

\appendix

\section{Algorithm for \textsc{LiteraryBigFive}}
\label{app:algorithm}

Algorithm~\ref{alg:literary_bigfive} presents the overall procedure of \textsc{LiteraryBigFive}, including the construction of interpretable literary axes, the localization of target authorial coordinates, and the adaptive steering of generation based on the style gap between the current hidden state and the target coordinates.

\begin{algorithm}[h]
\caption{\textsc{LiteraryBigFive} Framework.}
\label{alg:literary_bigfive}
\footnotesize
\begin{algorithmic}[1]
\Require LLM $M$, paired anchor passages $\mathcal{D}$, target passages $\{x_{b,i}\}_{i=1}^{m}$, input $x$, layers $\mathcal{L}$, strength $\lambda$.
\Ensure Stylized output $\hat{x}$.

\Statex \textbf{Offline: Literary space construction}
\For{each dimension $k\in\{1,\dots,5\}$ and layer $\ell\in\mathcal{L}$}
    \State Compute contrast vectors 
    $\boldsymbol{\delta}^{\ell}_{i}=a^{\ell}(x^{-}_{i}\oplus x^{+}_{i})-a^{\ell}(x^{-}_{i}\oplus x^{-}_{i})$.
    \State Obtain raw axis 
    $\tilde{\mathbf{v}}^{\ell}_{k}\leftarrow \frac{1}{N}\sum_{i=1}^{N}\boldsymbol{\delta}^{\ell}_{i}$.
\EndFor
\For{each layer $\ell\in\mathcal{L}$}
    \State Stack $\tilde{\mathbf{V}}^{\ell}=[\tilde{\mathbf{v}}^{\ell}_{1},\dots,\tilde{\mathbf{v}}^{\ell}_{5}]$ and perform SVD.
    \State Extract shared expressiveness axis $\mathbf{v}^{\ell}_{O}$.
    \State Remove $\mathbf{v}^{\ell}_{O}$ from each raw axis to obtain refined axes $\mathbf{V}^{\ell}$ and magnitudes $\boldsymbol{\rho}^{\ell}$.
\EndFor

\Statex \textbf{Online: Target localization}
\For{each layer $\ell\in\mathcal{L}$}
    \State $\mathbf{s}^{\ell}_{b}\leftarrow
    \frac{1}{m}\sum_{i=1}^{m}{\mathbf{V}^{\ell}}^{\top}\mathrm{norm}(a^{\ell}(x_{b,i}))$.
    \State $s^{\ell}_{O,b}\leftarrow
    \frac{1}{m}\sum_{i=1}^{m}
    \langle \mathrm{norm}(a^{\ell}(x_{b,i})),\mathbf{v}^{\ell}_{O}\rangle$.
\EndFor

\Statex \textbf{Online: Interpretable steering}
\For{each generated token $t$ and layer $\ell\in\mathcal{L}$}
    \State $\mathbf{s}^{\ell}_{t}\leftarrow {\mathbf{V}^{\ell}}^{\top}\mathrm{norm}(\mathbf{h}^{\ell}_{t})$,
    \quad
    $s^{\ell}_{O,t}\leftarrow
    \langle \mathrm{norm}(\mathbf{h}^{\ell}_{t}),\mathbf{v}^{\ell}_{O}\rangle$.
    \State $\boldsymbol{\alpha}^{\ell}_{t}\leftarrow
    \lambda\boldsymbol{\rho}^{\ell}\odot(\mathbf{s}^{\ell}_{b}-\mathbf{s}^{\ell}_{t})$,
    \quad
    $\alpha^{\ell}_{O,t}\leftarrow
    \lambda\rho^{\ell}_{O}(s^{\ell}_{O,b}-s^{\ell}_{O,t})$.
    \State ${\mathbf{h}^{\ell}_{t}}'\leftarrow
    \mathbf{h}^{\ell}_{t}
    +\mathbf{V}^{\ell}\boldsymbol{\alpha}^{\ell}_{t}
    +\alpha^{\ell}_{O,t}\mathbf{v}^{\ell}_{O}$.
\EndFor
\State \Return generated output $\hat{x}$.
\end{algorithmic}

\end{algorithm}

\begin{table*}[h]
\begin{small}
\begin{center}
\resizebox{\linewidth}{!}{
\begin{tabular}{l|cccc|cccc}
\toprule
\multirow{2}{*}{\textbf{Method}} 
& \multicolumn{4}{c|}{\textit{Reflections on the Revolution in France}} 
& \multicolumn{4}{c}{\textit{1984}} \\
& ROUGE-1 & ROUGE-L & SIM & GPT-4 
& ROUGE-1 & ROUGE-L & SIM & GPT-4 \\
\midrule
\textup{Few-shot} & 31.9 & 21.3 & 85.9 & 44.2 & 32.1 & 22.6 & 85.6 & 41.6 \\
\textup{LLM-Steer} & 39.5 & 26.2 & 89.3 & 41.8 & 44.1 & 31.4 & 88.6 & 39.8 \\
\textup{LoRA} & 42.2 & 35.7 & 90.0 & 46.5 & 62.5 & 56.2 & 94.8 & 52.4 \\
\textup{ICV} & 36.7 & 24.6 & 89.5 & 33.6 & 49.6 & 41.8 & 86.8 & 34.9 \\
\textup{Mean-Centering} & 50.6 & 40.2 & 94.0 & 47.4 & 56.2 & 48.3 & 92.9 & 47.1 \\
\textup{CAA} & 45.8 & 34.6 & 91.6 & 39.8 & 48.9 & 40.3 & 89.6 & 41.2 \\
\textup{RepE} & 38.8 & 30.2 & 86.5 & 43.1 & 42.2 & 35.4 & 86.3 & 45.6 \\
\rowcolor{green!10}\textsc{LiteraryBigFive} 
& \textbf{51.5} & \textbf{41.8} & \textbf{94.1} & \textbf{49.3} 
& \textbf{63.9} & \textbf{57.8} & \textbf{97.4} & \textbf{54.1} \\
\midrule
\multirow{2}{*}{\textbf{Method}} 
& \multicolumn{4}{c|}{\textit{Kidnapped}} 
& \multicolumn{4}{c}{\textit{Pride and Prejudice}} \\
& ROUGE-1 & ROUGE-L & SIM & GPT-4 
& ROUGE-1 & ROUGE-L & SIM & GPT-4 \\
\midrule
\textup{Few-shot} & 35.4 & 25.1 & 85.9 & 38.7 & 35.2 & 24.0 & 87.8 & 35.9 \\
\textup{LLM-Steer} & 44.4 & 31.6 & 88.5 & 40.8 & 41.6 & 28.2 & 89.4 & 37.2 \\
\textup{LoRA} & 59.1 & 52.5 & 96.5 & 50.6 & 56.2 & 47.5 & 95.7 & 54.3 \\
\textup{ICV} & 39.8 & 30.6 & 91.4 & 34.2 & 33.8 & 23.3 & 88.6 & 24.8 \\
\textup{Mean-Centering} & 51.0 & 42.3 & 92.5 & 52.1 & 49.1 & 38.1 & 91.7 & 45.0 \\
\textup{CAA} & 47.7 & 38.6 & 90.1 & 53.0 & 45.3 & 35.0 & 90.1 & 42.1 \\
\textup{RepE} & 49.2 & 40.7 & 90.2 & 56.4 & 44.1 & 33.8 & 87.9 & 49.5 \\
\rowcolor{green!10}\textsc{LiteraryBigFive} 
& \textbf{62.0} & \textbf{55.3} & \textbf{97.6} & \textbf{57.2} 
& \textbf{59.9} & \textbf{50.8} & \textbf{97.8} & \textbf{56.0} \\
\bottomrule
\end{tabular}
}
\end{center}
\end{small}
\vspace{-0.1in}
\caption{Experimental results on Qwen2.5-3B-Instruct. For all metrics, higher scores indicate better performance. The best-performing methods are highlighted in \textbf{bold}, and all results are scaled to 0--100.}
\vspace{-0.1in}
\vspace{-4mm}
\label{tab:qwen_backbone}
\end{table*}

\section{Generalization to Other Backbones}
\label{app:backbone_generalization}

To further evaluate the cross-model generalizability of \textsc{LiteraryBigFive}, we additionally conduct experiments on Qwen2.5-3B-Instruct~\cite{qwen2.5}. 
As shown in Table~\ref{tab:qwen_backbone}, \textsc{LiteraryBigFive} achieves the best performance across all four books and all evaluation metrics. 
These results suggest that the effectiveness of \textsc{LiteraryBigFive} is not tied to a specific backbone, and can extend across different model series and scales.

\section{Dataset Details}
\label{app:dataset}
\subsection{Dataset Construction}

Guided by the defined five dimensions, we curate an author-personalization dataset from English literary classics in the open-access {\href{https://www.gutenberg.org/}{\faExternalLink}}Gutenberg Library, which hosts over 75{,}000 ebooks.
The texts are freely available through Project Gutenberg, and we follow its Terms of Use and Project Gutenberg License for data access and redistribution.

To construct the \textsc{LiteraryBigFive} axes, we select 10 English literary classics, each authored by a distinct well-known writer, with details provided in Appendix~\ref{app:anchor_writers}.

For evaluation, we further curate four held-out books with distinct authorial expressions: \textit{Reflections on the Revolution in France} by Edmund Burke, \textit{1984} by George Orwell, \textit{Kidnapped} by R. L. Stevenson, and \textit{Pride and Prejudice} by Jane Austen.
Each book possesses a distinct authorial expression, allowing us to evaluate \textsc{LiteraryBigFive} across diverse writing patterns.

Due to formatting and compilation artifacts in the raw Gutenberg files which may distort analysis, we implement a cleaning pipeline to ensure the dataset focuses solely on literary content rather than formatting artifacts.
Specifically, we remove indentation symbols not present in the original texts and delete lines consisting of repeated ``='' symbols that lack semantic value.
Regarding line segmentation, we delete isolated line breaks used for visual alignment and reduce multiple consecutive line breaks to correctly preserve paragraph boundaries.
We also filter out unrelated segments, such as compiler contact information and hyperlinks.
Following this preprocessing, we segment the clean texts into passages with a length constraint of 120 to 400 tokens.
After preprocessing and segmentation, the axis-construction corpus comprises 1,322 passages spanning 12,741 sentences, while the held-out evaluation set comprises 590 passage-level samples totaling 5,716 sentences.

To construct authorial--neutral passage pairs, we rewrite each author-written passage $x^{+}$ into a neutralized version $x^{-}$ using an LLM, where the rewrite strips authorial cues while preserving the original meaning~\cite{ma2025dressing,zhang2025personalized}.
This pairing isolates authorial traits from content differences: $x^{+}$ and $x^{-}$ hold the same meaning, but only $x^{+}$ carries the author's voice.
Concretely, we prompt GPT-4~\cite{openai2024gpt4technicalreport} to suppress authorial cues across the five dimensions defined above without altering the core content, the prompt is provided in Appendix~\ref{app:neutralization}.
At evaluation time, the neutralized passage $x^{-}$ is used as input, and the original author-written passage $x^{+}$ serves as the target reference.

To verify data quality, we conduct an empirical analysis on 100 randomly selected passage pairs after preprocessing and neutralization. 
Specifically, we check whether formatting artifacts have been removed, whether the neutralized passage retains the core meaning of the original, and whether distinctive authorial expressions are sufficiently suppressed. 
Overall, the inspected pairs appear clean and suitable for both axis construction and evaluation.
This suggests that our pipeline removes formatting noise while preserving semantic content effectively, providing a usable contrast for extracting authorial directions.

\subsection{Anchor Writers and Books}
\label{app:anchor_writers}

To construct the \textsc{LiteraryBigFive} space, 
we selected representative ``anchor'' literary works for each dimension, whose writing patterns are representative of the corresponding dimension and have been discussed in prior literary and linguistic analysis.
The operational definitions and corresponding anchor books used to instantiate the positive direction of each axis are as follows:

\begin{itemize}

\item \textbf{Analyticity.} 
    This dimension focuses on logical reasoning and propositional density. Following Biber’s Multidimensional Analysis~\cite{biber1995dimensions}, high analyticity is marked by a high frequency of \textit{abstract nouns} and \textit{logical connectors} (e.g., causal and conditional links), which facilitate complex information integration.
    The selected books are as follows:
    \begin{itemize}
        \item \textit{The Sacred Wood} by T. S. Eliot 
        \item \textit{The Problems of Philosophy} by Bertrand Russell
    \end{itemize}
    \textbf{Rationale:} These works represent a logic-driven style that values intellectual clarity. 
    Both Eliot and Russell provide representative examples of argument-driven prose, where the writing is organized around conceptual development and logical progression~\cite{eliot2024sacred}.

    \item \textbf{Ornateness.} 
    This dimension represents aesthetic richness. It is characterized by \textit{high vocabulary diversity} and \textit{complex sentence structures}, particularly through the frequent use of descriptive phrases and extra details attached to nouns to create vivid imagery~\cite{lanham1991handlist}. 
The selected works are as follows:
    \begin{itemize}
        \item \textit{Sartor Resartus} by Thomas Carlyle
        \item \textit{The Renaissance} by Walter Pater
    \end{itemize}
    \textbf{Rationale:} Carlyle’s distinctive and elaborate prose style is closely tied to the complex philosophical ideas it conveys~\cite{levine1968carlylese}, while Pater’s work is closely associated with the Aesthetic Movement, using rhythmic and highly decorated sentences to elevate the sensory experience of the reader~\cite{pater2023renaissance}.

    \item \textbf{Narrativity.} 
    This dimension captures event-driven storytelling. Following established narrative theory~\cite{labov1972language}, high narrativity is identified by the frequent use of \textit{action verbs} and \textit{time markers} (e.g., ``then,'' ``afterward'') that move the plot forward in a clear sequence.
    The selected works are as follows:
    \begin{itemize}
        \item \textit{Robinson Crusoe} by Daniel Defoe \item \textit{The Call of the Wild} by Jack London
    \end{itemize}
    \textbf{Rationale:} These texts provide representative examples of linear, event-driven storytelling. Defoe’s \textit{Crusoe} is widely discussed for its step-by-step account of physical actions~\cite{watt1957rise}, while London’s direct and action-focused prose offers another anchor for narrative progression.

    \item \textbf{Emotionality.} 
    This axis measures the intensity of the characters' internal feelings and psychological states. It is characterized by the use of \textit{emotive adjectives}, \textit{exclamations}, and verbs related to internal thoughts, reflecting the ``inward turn'' of the novel~\cite{auerbach2013mimesis}. 
The selected works are as follows:
\begin{itemize}
    \item \textit{Mrs. Dalloway} and \textit{To the Lighthouse} by Virginia Woolf
    \item \textit{Sons and Lovers} and \textit{Women in Love} by D. H. Lawrence
\end{itemize}
\textbf{Rationale:} These works prioritize affective subjective experience over external plot. Woolf’s novels are famous for capturing the fluid "stream of consciousness"~\cite{auerbach2013mimesis}, while Lawrence’s novels explore the raw, deep-seated emotional and psychological tensions between individuals~\cite{niven1978dh}.

\item \textbf{Classicism.} 
This dimension captures formal and balanced prose patterns associated with 18th- and 19th-century English writing.
This period is chosen because it reflects relatively standardized prose conventions, including shared expectations about structure and decorum before many 20th-century experimental writing practices~\cite{mckeon2002origins}.
The selected works are as follows:
\begin{itemize}
    \item \textit{The Rambler} by Samuel Johnson
    \item \textit{The Spectator} by Addison and Steele
\end{itemize}
\textbf{Rationale:} 
These authors are central figures in the ``Golden Age'' of English essay writing.
Johnson's work exemplifies Neoclassical balance and symmetry~\cite{wimsatt1941prose}, while the essays in \textit{The Spectator} helped popularize a formal, polite, and standardized prose style.

\end{itemize}

\section{Effectiveness of Axis Decomposition}

\begin{figure}[H]
\centering
\includegraphics[width=\columnwidth]{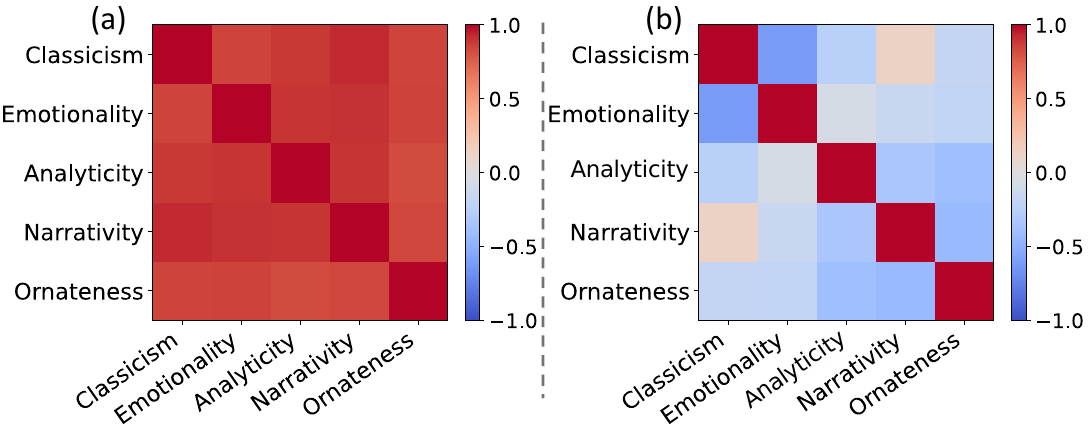}  
\caption{
Pairwise cosine similarity heatmaps of the five axes before and after decomposition.
(a) Before decomposition, the raw axes are strongly correlated.
(b) After decomposition, cross-axis similarity is substantially reduced.
Results shown for LLaMA2-7B-Chat.
}
\vspace{-2mm}
  \label{fig:pre_post_decomp_heatmap}
\end{figure}

To directly verify the effect of the decomposition step, Figure~\ref{fig:pre_post_decomp_heatmap} compares the pairwise cosine similarity heatmaps of the five axes before and after decomposition. 
Before decomposition, the raw axes exhibit uniformly high positive cross-axis similarity, indicating a substantial shared component, which we term the overall expressiveness direction.

After decomposition, the cross-axis similarities are markedly reduced, with the mean absolute off-diagonal cosine decreasing from 0.87 to 0.27. This substantial drop provides direct evidence in Section~\ref{sec:refinement} that the decomposition effectively removes the shared global trend among the raw axes, thereby yielding more disentangled and dimension-specific directions for stable multi-axis composition.

\section{Coordinate Calibration}
\label{app:calibration}
Raw per-axis coordinates can differ in dynamic range across dimensions, so we calibrate them using the \emph{all anchor passages} employed to construct each axis (i.e., all passages from the representative books for that dimension).
For the $k$-th axis, let $\mathcal{A}_k$ be its anchor corpus, we compute layer-averaged projection on axis $k$ for each passage $x_i \in \mathcal{A}_k$:
\[
s(x_i;k)\;=\;\textstyle  \frac{1}{|\mathcal{L}|}\sum_{\ell\in\mathcal{L}}
\langle \widehat{a}^{\ell}(x_i),\,\mathbf{v}^{\ell}_{k}\rangle,
\]
and collect the projections of all passages
$\mathcal{P}_k=\{s(x_i;k):x_i\in\mathcal{A}_k\}$.
We then set an axis-specific scale $\alpha_k$ from this distribution $\mathcal{P}_k$ to make coordinates comparable across dimensions. Concretely, we use the 95th percentile of $|p|$, which is a standard robust-scaling choice that limits the influence of outliers, avoids saturating typical books at the bounds, and remains stable as the anchor corpus grows:
\[
\alpha_k \;=\; \operatorname{quantile}_{0.95}\big(\{|p|:p\in\mathcal{P}_k\}\big).
\]
Given a book’s layer-averaged raw scores $\mathbf{s}_b\in\mathbb{R}^5$ from \S~\ref{localization}, its calibrated coordinate on axis $k$ is
\[
z_{b,k}\;=\;\mathrm{clip}\!\left(\frac{s_{b,k}}{\alpha_k},\, -1,\, 1\right),
\]
combining every dimension together forms a five-dimensional score vector $\mathbf{z}_b \in [-1, 1]^5$.
Building upon this, we map to $[0,100]$ for visualization in radar plots:
\[
R_{b,k}\;=\;50\,(1+z_{b,k}), 
\quad \mathbf{R}_b=(R_{b,1},\dots,R_{b,5}).
\]

\section{Detailed Evaluation Scores}

In this section, we report the complete GPT-4 and human evaluation scores across the two dimensions, namely Semantic Fidelity (SF) and Authorial Adherence (AA). Table~\ref{tab:subjective_res_weighted_overall} presents the average performance, providing the numerical data visualized in Figure~\ref{fig:sasf}, while Table~\ref{tab:subjective_res} provides a detailed breakdown for each of the four books.

\begin{table*}[t]
\centering
\small 
\setlength{\tabcolsep}{14pt} 
\begin{tabular}{lcccccccc} 
\toprule
\multirow{2.5}{*}{\textbf{Method}} & \multicolumn{4}{c}{\textit{Reflections on the Revolution in France}} & \multicolumn{4}{c}{\textit{1984}} \\
\cmidrule(lr){2-5} \cmidrule(lr){6-9} 
 & \multicolumn{2}{c}{GPT-4} & \multicolumn{2}{c}{Human} & \multicolumn{2}{c}{GPT-4} & \multicolumn{2}{c}{Human} \\
 & SF & AA & SF & AA & SF & AA & SF & AA \\
\midrule
\textup{Few-shot}       & 72.2 & 57.0 & 50.0 & 35.0 & 77.3 & 62.1 & 73.9 & 35.1 \\
\textup{LLM-Steer}      & 76.3 & 54.4 & 74.5 & 46.0 & 79.0 & 58.7 & 72.7 & 35.1 \\
\textup{LoRA} & 68.4 & 53.2 & 52.8 & 46.2 & 75.1 & 63.4 & 60.6 & 53.8 \\
\textup{ICV}           & 77.4 & 53.4 & 72.3 & 53.2 & 84.1 & 62.2 & 83.6 & 59.3 \\
\textup{Mean-Centering} & 80.4 & 55.4 & 78.2 & 53.5 & 84.6 & 63.1 & 85.7 & 62.0 \\
\textup{CAA}            & 67.5 & 51.9 & 59.5 & 32.0 & 64.0 & 51.1 & 56.1 & 37.3 \\
\textup{RepE}           & \textbf{81.0} & 56.1 & 77.7 & 53.0 & 84.9 & 62.0 & 83.5 & 58.2 \\
\rowcolor{green!10}\textsc{LiteraryBigFive} & 80.8 & \textbf{58.1} & \textbf{81.8} & \textbf{56.5} & \textbf{85.5} & \textbf{65.0} & \textbf{86.5} & \textbf{64.1} \\
\midrule
\multirow{2.5}{*}{\textbf{Method}} & \multicolumn{4}{c}{\textit{Kidnapped}} & \multicolumn{4}{c}{\textit{Pride and Prejudice}} \\
\cmidrule(lr){2-5} \cmidrule(lr){6-9}
 & \multicolumn{2}{c}{GPT-4} & \multicolumn{2}{c}{Human} & \multicolumn{2}{c}{GPT-4} & \multicolumn{2}{c}{Human} \\
 & SF & AA & SF & AA & SF & AA & SF & AA \\
\midrule
\textup{Few-shot}       & 67.5 & 49.5 & 45.5 & 48.8 & 61.7 & 43.4 & 54.8 & 42.9 \\
\textup{LLM-Steer}      & 71.2 & 46.3 & 73.3 & 40.8 & 66.5 & 43.9 & 61.9 & 41.0 \\
\textup{LoRA} & 65.3 & 52.1 & 60.1 & 52.6 & 62.4 & 49.8 & 57.2 & 50.2 \\
\textup{ICV}           & 79.1 & 50.8 & 85.0 & 45.9 & 74.1 & 48.9 & 74.4 & 51.5 \\
\textup{Mean-Centering} & 80.3 & 52.8 & 78.2 & 45.3 & 77.0 & 51.0 & 77.8 & 55.6 \\
\textup{CAA}            & 64.9 & 47.0 & 50.8 & 31.2 & 59.8 & 42.7 & 59.5 & 38.6 \\
\textup{RepE}           & 80.4 & 51.4 & 84.2 & 52.3 & 76.3 & 50.7 & 82.5 & 53.1 \\
\rowcolor{green!10}\textsc{LiteraryBigFive} & \textbf{81.6} & \textbf{54.1} & \textbf{89.7} & \textbf{57.8} & \textbf{78.5} & \textbf{53.3} & \textbf{80.5} & \textbf{58.5} \\
\bottomrule
\end{tabular}
\caption{GPT-4 and Human evaluation across four books on two dimensions: Semantic Fidelity (SF) and Authorial Adherence (AA). Best results are \textbf{bolded}, all results are scaled to a 0--100 scale.}
\vspace{-2mm}
\label{tab:subjective_res}
\end{table*}

\section{Authorial Coordinate Scores}

\label{app:style_score}

We provide the detailed authorial scores in Figure~\ref{fig:radar}, as shown in Table~\ref{tab:style_multi_model}.

\begin{figure*}[t]
  \includegraphics[width=\linewidth]{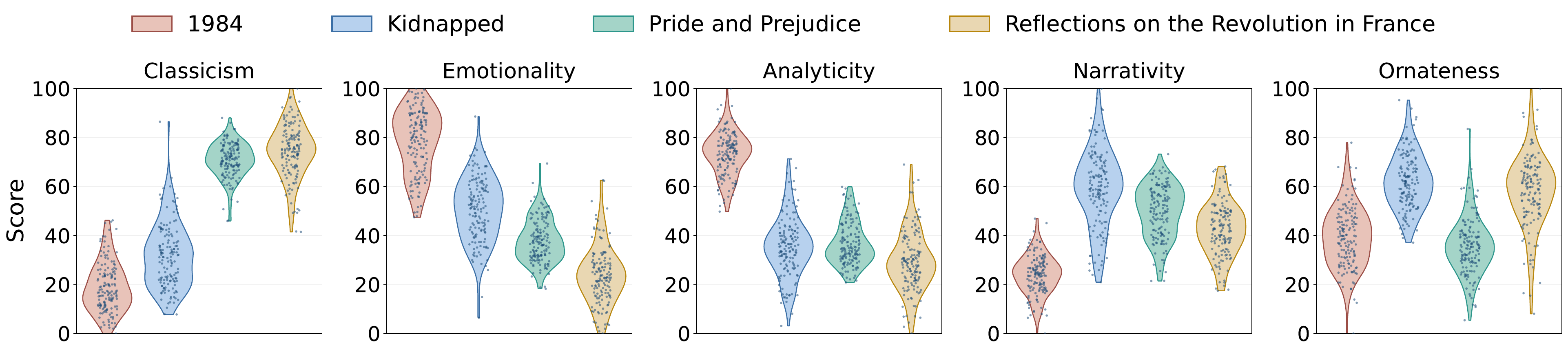} 
  \caption {Per-book distributions across the five stylistic dimensions. The clear separation between books matches their known literary characteristics, demonstrating the framework's effectiveness.}
  \label{fig:distribution}
\vspace{-4mm}

\end{figure*}

To better show how these stylistic dimensions separate by book, we plot the score distributions for each author-written passage from our test set.
As shown in Figure~\ref{fig:distribution}, the results are very consistent with known writing patterns of the selected books. Orwell's \textit{1984} has a high level of \textit{Analyticity}, which fits with its focus on complex political and social critique. In contrast, Burke's \textit{Reflections} shows the highest scores for \textit{Classicism} and \textit{Ornateness}, as expected for formal, highly-stylized 18th-century work. \textit{Kidnapped} stands out for \textit{Narrativity}, reflecting its narrative-driven adventure story style. These clear, separated distributions prove that our \textsc{LiteraryBigFive} framework can accurately capture and distinguish different author characteristics.

\begin{table*}[t]
\begin{small}
\begin{center}
\resizebox{\linewidth}{!}{
\begin{tabular}{l|ccccc|ccccc}
\toprule
\multirow{2}{*}{\textbf{Model}} 
& \multicolumn{5}{c|}{\textit{Reflections on the Revolution in France}} 
& \multicolumn{5}{c}{\textit{1984}} \\
& Classicism & Emotionality & Analyticity & Narrativity & Ornateness
& Classicism & Emotionality & Analyticity & Narrativity & Ornateness \\
\midrule
\rowcolor{green!10}\textsc{LiteraryBigFive} & 75.7 & 12.1 & 25.7 & 49.4 & 60.4 
                & 30.3 & 69.5 & 72.9 & 38.5 & 50.6 \\
GPT-5           & 82.0 & 25.0 & 18.0 & 42.0 & 72.0
                & 30.0 & 75.0 & 78.0 & 55.0 & 40.0 \\
Gemini 3        & 78.5 & 18.0 & 32.0 & 44.0 & 65.0
                & 28.0 & 65.0 & 76.0 & 45.0 & 42.0 \\
Claude-Sonnet-3.5 & 77.0 & 20.0 & 28.0 & 45.0 & 65.0
                & 29.0 & 70.0 & 75.0 & 46.0 & 44.0 \\
\midrule
\multirow{2}{*}{\textbf{Model}} 
& \multicolumn{5}{c|}{\textit{Kidnapped}} 
& \multicolumn{5}{c}{\textit{Pride and Prejudice}} \\
& Classicism & Emotionality & Analyticity & Narrativity & Ornateness
& Classicism & Emotionality & Analyticity & Narrativity & Ornateness \\
\midrule
\rowcolor{green!10}\textsc{LiteraryBigFive} & 39.0 & 44.7 & 38.9 & 61.9 & 59.8
                & 66.6 & 30.3 & 38.9 & 49.7 & 55.6 \\
GPT-5           & 45.0 & 45.0 & 50.0 & 78.0 & 48.0
                & 72.0 & 38.0 & 52.0 & 62.0 & 48.0 \\
Gemini 3        & 42.0 & 52.0 & 35.0 & 70.0 & 62.0
                & 70.0 & 35.0 & 48.0 & 53.0 & 52.0 \\
Claude-Sonnet-3.5 & 42.0 & 47.0 & 41.0 & 70.0 & 56.0
                  & 69.0 & 34.0 & 46.0 & 55.0 & 52.0 \\
\bottomrule
\end{tabular}
}
\end{center}
\end{small}
\caption{Comparison of \textsc{LiteraryBigFive} dimension scores across models on four books. Higher indicates stronger presence of the corresponding attribute.}
\vspace{-3mm}

\label{tab:style_multi_model}
\end{table*}

\begin{table}[t]
\centering
\small
\setlength{\tabcolsep}{14pt}
\resizebox{\linewidth}{!}{%
\begin{tabular}{lcccc}
\toprule
\textbf{Method} & \multicolumn{2}{c}{GPT-4} & \multicolumn{2}{c}{Human} \\
\cmidrule(lr){2-3}\cmidrule(lr){4-5}
& SF & AA & SF & AA \\
\midrule
\textup{Few-shot}        & 69.7 & 53.0 & 56.2 & 40.4 \\
\textup{LLM-Steer}       & 73.2 & 50.8 & 70.5 & 40.7 \\
\textup{LoRA} & 67.8 & 54.6 & 57.7 & 50.7 \\
\textup{ICV}             & 78.7 & 53.9 & 78.8 & 52.5 \\
\textup{Mean-Centering}  & 80.6 & 55.6 & 80.0 & 54.2 \\
\textup{CAA}             & 64.0 & 48.2 & 56.5 & 34.8 \\
\textup{RepE}            & 80.6 & 55.1 & 82.0 & 54.2 \\
\rowcolor{green!10}\textsc{LiteraryBigFive} & \textbf{81.6} & \textbf{57.7} & \textbf{84.6} & \textbf{59.3} \\
\bottomrule
\end{tabular}%
}
\caption{Performance comparison of different methods on two dimensions: Semantic Fidelity (SF) and Authorial Adherence (AA). Best results are \textbf{bolded}, all results are scaled to 0--100.}
\vspace{-2mm}
\label{tab:subjective_res_weighted_overall}
\end{table}

\begin{figure*}[t]
  \includegraphics[width=\linewidth]{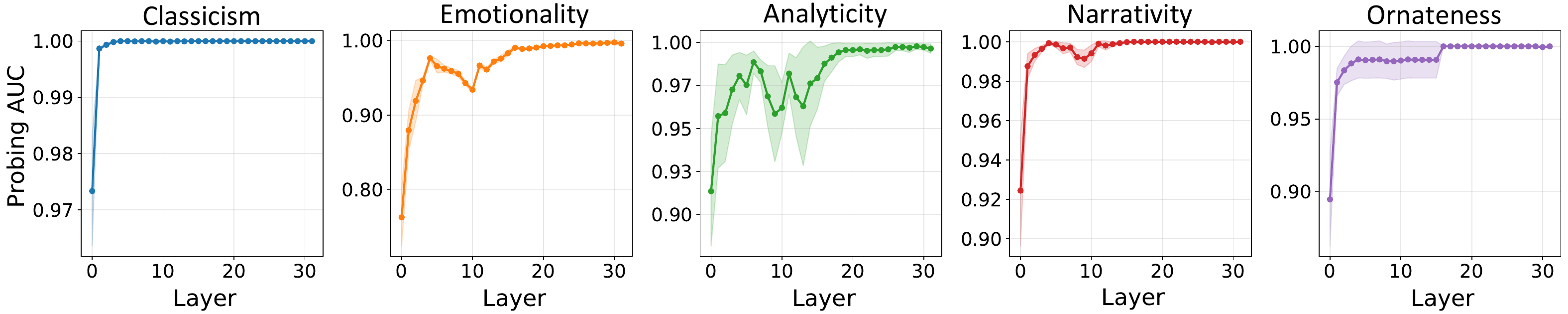} 
  \caption {Layer-wise linear probing performance (AUC) across the five \textsc{LiteraryBigFive} dimensions. The results reveal a hierarchical encoding mechanism: surface-level attributes (e.g., \textit{Classicism}, \textit{Narrativity}) saturate rapidly in early layers, whereas complex semantic attributes (e.g., \textit{Emotionality}, \textit{Analyticity}) require deeper processing to reach maximal separability.}
  \label{fig:probing}
\vspace{-3mm}

\end{figure*}

\section{Efficiency Comparison}

We analyze the computational efficiency of \textsc{LiteraryBigFive} from both theoretical and empirical perspectives, as summarized in Table~\ref{tab:latency_comparison}.

\begin{table}[h]
\centering
\small
\begin{tabular}{lcc}
\toprule
\textbf{Method} & \textbf{Complexity} & \textbf{Latency} \\
\midrule
\multicolumn{3}{l}{\textit{Fine-tuning Methods}} \\
\quad LoRA & $O(r \cdot d)$ & 23.50 \\
\quad LLM-Steer & $O(d^2)$ & 19.92 \\
\midrule
\multicolumn{3}{l}{\textit{Activation Steering Methods}} \\
\quad ICV & $O(d)$ & 19.49 \\
\quad Mean-Centering & $O(d)$ & 18.93 \\
\quad RepE & $O(d)$ & 19.29 \\
\quad CAA & $O(d)$ & 19.02 \\
\rowcolor{green!10} \textsc{LiteraryBigFive} & $O(K \cdot d)$ & 19.88 \\
\bottomrule
\end{tabular}
\caption{Efficiency comparison. We report the theoretical Computational Complexity per token and the measured Inference Latency (ms/token).}
\label{tab:latency_comparison}
\vspace{-2mm}
\end{table}

\textbf{Theoretical Complexity.}
Our method maintains a linear computational complexity of $O(K \cdot d)$ per token, where $K$ is the number of vectors used to intervene the models (here $K=6$) and $d$ is the hidden dimension. 
This represents a significant theoretical advantage over fine-tuning methods like LLM-Steer~\cite{han2024word}, which require a dense matrix multiplication with quadratic complexity $O(d^2)$. 
Even compared to parameter-efficient methods such as unmerged LoRA~\cite{hu2022lora} with complexity $O(r \cdot d)$ (where $r$ is the rank, in our settings $r=8$), our approach remains more efficient as $K \le r \ll d$. Given that $6 \le 8 \ll 4096$ for Llama-2-7B-Chat, the theoretical FLOPs required by our steering mechanism are orders of magnitude lower than fine-tuning and more streamlined than LoRA configurations.

\textbf{Inference Latency.}
To evaluate real performance, we measured the average inference latency (ms/token) over 100 generated cases in our test set. 
As shown in Table~\ref{tab:latency_comparison}, static vector-based baselines (e.g., Mean-Centering, CAA) exhibit the lowest latency ($\sim$18.9--19.0 ms/token) since they apply a fixed bias.
In contrast, the training-based LoRA baseline incurs higher latency (23.50 ms/token) due to the additional low-rank adapter computation.
Despite the computational overhead of calculating projections and style gaps along $K$ axes for dynamic adaptation, \textsc{LiteraryBigFive} records a latency of 19.88 ms/token.
This corresponds to a marginal overhead of less than 1.0 ms compared to the fastest static baseline (Mean-Centering, 18.93 ms) and is effectively equivalent to LLM-Steer (19.92 ms).
These results demonstrate that our method's dynamic control comes at a practically negligible cost, remaining highly efficient for real-time generation while offering the unique capability of disentangled, interpretable personalized steering that static vector addition cannot achieve.

\section{\textsc{BigFive} Dimension Vector Analysis}

To investigate where and how the \textsc{LiteraryBigFive} dimensions are encoded within the model's internal representations, we conduct a layer-wise linear probing analysis.
Specifically, for each dimension, we train a logistic regression classifier on the hidden states $\mathbf{h}^\ell$ extracted from each layer $\ell$ to distinguish between texts exhibiting high versus low intensity along that dimension.
Figure~\ref{fig:probing} illustrates the probing AUC trajectories across model layers, revealing two critical insights into how these dimensions are represented in the model.

\paragraph{Universal Dimension Encodability.}
First, we observe that the model achieves high classification performance
(AUC $> 0.90$) across all five dimensions.
This indicates that dimension-level information is not an abstract external label, but is robustly embedded within the LLM's latent space.
Even without explicit supervision during pre-training, the model learns representations that distinguish these dimension-specific patterns, validating the probing-based foundation of our steering approach.

\paragraph{Hierarchical Encoding of Each Dimension.}
Crucially, our fine-grained analysis reveals a clear layer-wise hierarchy regarding when different dimensions become linearly separable.
While all dimensions are eventually encoded, they do so at different depths within the network:

\begin{itemize}
    \item \textbf{Surface-Level Dimensions (Classicism, Narrativity):}
    As shown in the plots for \textit{Classicism} and \textit{Narrativity}, the AUC scores saturate rapidly, reaching near-perfect performance within the first few layers (Layers 0--5).
    This suggests that these dimensions are closely associated with lexical markers (e.g., archaic function words) or shallow syntactic patterns (e.g., verb and event distributions), which are captured early in the bottom-up processing.

    \item \textbf{Semantic-Level Dimensions (Analyticity, Emotionality, Ornateness):}
    In contrast, dimensions such as \textit{Analyticity}, \textit{Emotionality}, and \textit{Ornateness} exhibit a more gradual ascent in AUC, peaking only in the middle-to-late layers (Layers 15--25).
    \textit{Analyticity}, in particular, shows higher variance in lower layers, indicating that its reliable representation requires compositional reasoning and long-range contextual integration.
\end{itemize}

Overall, these results indicate that while shallow layers encode surface-level lexical and structural patterns, the representation of more abstract reasoning processes and affective nuances relies on the deeper abstraction capabilities of the network.

\begin{table*}[h]
\centering
\resizebox{\linewidth}{!}{%
\begin{tabular}{llccccc}
\toprule
\textbf{Author} & \textbf{Work} & \textbf{Classicism} & \textbf{Emotionality} & \textbf{Analyticity} & \textbf{Narrativity} & \textbf{Ornateness} \\ 
\midrule
Ernest Hemingway & \textit{The Old Man and the Sea}      & 24.0 & 81.1 & 65.1 & 45.4 & 51.0 \\
William Faulkner & \textit{The Sound and the Fury}      & 23.9 & 81.4 & 53.8 & 44.2 & 55.6 \\ 
\midrule
Francis Bacon    & \textit{The Essays}                  & 68.3 & 9.5  & 33.6 & 67.8 & 63.8 \\
Agatha Christie  & \textit{Murder on the Orient Express} & 31.3 & 57.4 & 61.5 & 34.0 & 62.8 \\ 
\midrule
John Henry Newman & \textit{Apologia Pro Vita Sua}      & 51.9 & 27.0 & 38.0 & 43.2 & 66.0 \\
Kazuo Ishiguro   & \textit{Never Let Me Go}             & 24.3 & 80.2 & 66.7 & 49.6 & 48.7 \\ 
\bottomrule
\end{tabular}
}
\caption{Style coordinates for additional canonical authors in the \textsc{LiteraryBigFive} space. Higher values indicate a stronger presence of the corresponding attribute.}
\label{tab:extended_coordinates}
\vspace{-1mm}
\end{table*}


\section{More Authorial Coordinates Analysis}
To further validate the robustness and discriminative ability of the \textsc{LiteraryBigFive} space across a broader spectrum of authors, we analyze six additional authors with distinct writing patterns.
Table~\ref{tab:extended_coordinates} presents their localized coordinates, demonstrating how the model situates diverse authorial patterns within our five-dimensional framework.

The model’s positioning aligns closely with established literary criticism. For instance, while Ernest Hemingway and William Faulkner both show high \textit{Emotionality}, they are separated by \textit{Ornateness} ($\Delta=4.6$). Hemingway’s lower score quantitatively reflects his ``Iceberg Theory,'' which favors a sparse, direct lexicon over decorative language~\cite{hemingway1999death}, whereas Faulkner’s higher score captures his famously multi-layered sentence structures~\cite{faulkner1956william}.
Similarly, the contrast between Francis Bacon and Agatha Christie highlights nuances in \textit{Narrativity}. Bacon’s high scores in \textit{Narrativity} (67.8) and \textit{Classicism} (68.3) reflect the 17th-century rhetorical tradition, where progression is driven by explicit logical steps~\cite{vickers1968francis}. 
Conversely, Christie’s lower \textit{Narrativity} (34.0) reflects a style that relies more on dialogue and internal deduction than on physical action. 
Finally, the model captures the historical shift from 19th-century eloquence to modern restraint. John Henry Newman’s high \textit{Ornateness} (66.0) is consistent with Victorian rhythmic and stylized prose~\cite{henkle1970boundaries}, while Kazuo Ishiguro’s lower score (48.7) and high \textit{Emotionality} (80.2) accurately represent his intentional use of "plainspoken" language to mask deep psychological tension~\cite{ishiguro2007art}.

\vspace{-1mm}
\section{Baseline Details}
\label{app:baseline}

In this section, we describe the baseline methods used in our experiments, categorized into prompting, fine-tuning, and activation steering.

First, we use few-shot prompting as a basic comparison.
Specifically, we prepend $k$ reference passages written by the target author to the input prompt.
This baseline evaluates how well the model can adapt its generation to an author’s writing characteristics purely through in-context examples, without modifying any internal parameters.
The specific prompts are listed in Appendix~\ref{app:fewshot}.

Second, for methods that require training, we adopt LLM-Steer~\cite{han2024word} and LoRA \cite{hu2022lora}.
Instead of retraining the entire model, LLM-Steer learns a lightweight linear transformation over word embeddings to align the generated text with the target author’s writing patterns, while LoRA injects trainable low-rank adapters into selected layers to achieve parameter-efficient style adaptation with a frozen backbone.

Finally, we compare our approach against four representative activation steering methods that intervene directly in the model's hidden states:
(1) ICV~\cite{liu2024context}, which extracts intervention vectors from few reference examples;
(2) Mean-Centering~\cite{jorgensen2023improving}, which computes a fixed direction by subtracting the average activations of neutral rewrites from those of the target author's expressions;
(3) CAA~\cite{rimsky2024steering}, which derives a steering direction by contrasting activations between author-written and neutral texts, {thereby covering the mean-difference steering formulation used by StyleVector for personalized text generation~\cite{zhang2025personalized};
and (4) RepE~\cite{zou2023transparency}, which identifies the principal direction of linguistic variation via PCA on contrastive activation pairs.

Since these baselines are typically designed for single-target steering and lack an inherent unified style space, we adapt them to our framework to ensure a fair comparison.
Specifically, for all methods except ICV, we aggregate the anchor books used to construct each style axis (see Section~\S\ref{space}) and utilize this full data for training, while ICV follows its standard setup.

\section{Implementation Details}

\label{app:implementation}

We compare \textsc{LiteraryBigFive} against several state-of-the-art steering and prompting baselines with specific hyperparameter configurations. 
For Few-shot prompting, we randomly sample 3 passages to guide generation.
For LLM-Steer, we use the learned transform with $\epsilon_0 = 1\times10^{-3}$ scaled by a factor of 6 (i.e., $\epsilon = 6\epsilon_0$).
For LoRA, we set the rank to 8 and train for 3 epochs.
We use a cosine learning-rate scheduler with a warm-up stage covering 10\% of total steps, with the maximum learning rate set to $5\times10^{-5}$.
The batch size is 2 with gradient accumulation of 16.
For methods involving activation steering, we adhere to the following configurations, selected by grid search:
(1) For ICV, we randomly sample 3 reference passages to extract vectors and apply the intervention across all layers except the first embedding layer, with a steering strength $\alpha=0.3$.
(2) Regarding Mean-Centering, the editing strength is set to $\alpha=1$ applied to layers $\ell \in \{22, 23, 25, 27, 29\}$.
(3) For CAA, we set the scaling $\alpha=1$ for layers $\ell \in \{23, 25, 27, 29\}$.
(4) For RepE, we configure $\alpha=6$ targeting layers $\ell \in \{22, 23, \dots, 29\}$.
Finally, for our proposed \textsc{LiteraryBigFive}, we apply steering vectors with a global strength of $\lambda=1$ across layers $\ell \in \{20, 24, 28\}$ and randomly sample 10 passages as reference.
To rigorously evaluate semantic preservation and eliminate metric instability caused by random sampling, we set the decoding temperature to 0 for all experiments to ensure deterministic outputs.

\section{Case Study}
\label{app:case_study}

We present representative examples from the test sets of each book. We highlight \good{desirable linguistic features} in green and \bad{hallucinations or flattening} in red, followed by a detailed analysis for each case.

\begin{table*}[t]
\centering
\small 
\renewcommand{\arraystretch}{1.1} 
\begin{tabularx}{\textwidth}{X X X}
\toprule
\multicolumn{1}{c}{\textbf{\textsc{LiteraryBigFive} (Ours)}} & 
\multicolumn{1}{c}{\textbf{RepE}} & 
\multicolumn{1}{c}{\textbf{Few-Shot}} \\
\midrule

\multicolumn{3}{p{0.96\textwidth}}{\textbf{Original Text 1:} ``The first, calling itself the \textbf{Constitutional Society}... What improvements they have had... \textbf{I cannot tell: But I never heard a man... speak a word in praise} of the greater part of the publications...''} \\
\midrule

The first group, known as the \good{Constitutional Society}... While it is possible that some of the books were sent to France... \good{I have never heard anyone praise} the publications circulated by this society... & 
The first group, known as the Constitutional Society... Some claim that books transported across the sea can improve... but \bad{this is unverified}. Despite good intentions... their publications \bad{have not been widely praised}... & 
The \bad{National Assembly}, composed of practitioners in the law, has been around for about seven or eight years. ... I have heard talk about the glories of the \bad{National Assembly}... \\

\multicolumn{3}{p{0.96\textwidth}}{\vspace{0.3em}\textit{\textbf{Analysis:}} Few-Shot suffers from severe hallucination, replacing ``Constitutional Society'' with \bad{``National Assembly''}. RepE accurately keeps the entity but \bad{flattens the style}: it converts the author's strong first-person voice (``I cannot tell'', ``I never heard'') into dry, objective summaries (``this is unverified'', ``have not been widely praised''). Our method preserves the \good{subjective, critical voice} essential to Burke's style.\vspace{0.3em}} \\
\midrule

\multicolumn{3}{p{0.96\textwidth}}{\textbf{Original Text 2:} ``Well! but these men were to be tempered... \textbf{Were they then to be awed} by the... dignity of an handful of country clowns...? \textbf{No!} both these descriptions were more formed to be overborne...''} \\
\midrule

The men were supposed to be influenced... \good{However, were they intimidated} by the authority and dignity of a few country people in the Assembly...? \good{No, both groups were more likely} to be influenced by the tactics of lawyers... & 
The men were supposed to be influenced... \bad{However, they were unlikely to be intimidated} by the authority and dignity of a few country people... The lawyers would likely govern the whole... & 
\bad{The clergy had much to do with these transactions. Their estates were bound to the last acre.} They were false to every other interest and held out the church lands as a security... \\

\multicolumn{3}{p{0.96\textwidth}}{\vspace{0.2em}\textit{\textbf{Analysis:}} Few-Shot retrieves \bad{completely irrelevant context} about ``the clergy'' from a different chapter. RepE correctly identifies the content but \bad{loses the rhetorical structure}, turning a dramatic Question-Answer sequence (``Were they...? No!'') into a flat statement (``they were unlikely to be''). Our method successfully retains the \good{rhetorical question and emphatic negation}, preserving the argumentative tension.} \\
\bottomrule
\end{tabularx}
\caption{Qualitative comparison on \textit{Reflections on the Revolution in France} by Edmund Burke. We highlight \good{desirable linguistic features} in green and \bad{hallucinations or flattening} in red. }
\label{tab:case_study}
\vspace{-0.1in}
\end{table*}

\begin{table*}[t]
\centering
\small 
\renewcommand{\arraystretch}{1.1} 
\begin{tabularx}{\textwidth}{X X X}
\toprule
\multicolumn{1}{c}{\textbf{\textsc{LiteraryBigFive} (Ours)}} & 
\multicolumn{1}{c}{\textbf{RepE}} & 
\multicolumn{1}{c}{\textbf{Few-Shot}} \\
\midrule

\multicolumn{3}{p{0.96\textwidth}}{\textit{\textbf{Original Passage 1:}} ``...the shouting, the singing, the banners, the posters, the films, the waxworks, the rolling of drums and squealing of trumpets, the tramp of marching feet... when the great orgasm was quivering to its climax...''} \\
\midrule

On the sixth day of Hate Week, \good{after the events such as processions, speeches, shouting, singing, banners, posters, films, waxworks, drums,} trumpets, marching feet, tank caterpillars, planes, and guns, when the \good{intense excitement was reaching its peak}... & 
On the sixth day of Hate Week, \bad{after the events such as processions, speeches,} shouting, singing, banners, posters, films, waxworks, drums, trumpets, marching feet, tank caterpillars, planes, and guns, \bad{when the intense excitement was reaching its peak}... & 
On the sixth day of Hate Week, after the events such as processions, speeches, shouting, singing, banners, posters, films, waxworks, drums, trumpets, marching feet, tank caterpillars, planes, and guns, \bad{after six days of this...} \\

\multicolumn{3}{p{0.96\textwidth}}{\vspace{0.3em}\textit{\textbf{Analysis:}} While RepE converts the vivid description into a mechanical list (highlighted in \bad{red}), losing the sensory overload, our method \good{preserves the density and rhythm} of the ``wall of sound'' effect essential to the scene's atmosphere.\vspace{0.3em}} \\
\midrule

\multicolumn{3}{p{0.96\textwidth}}{\textit{\textbf{Original Passage 2:}} ``The past, he reflected, had not merely been altered, it had been actually destroyed. For how could you establish even the most obvious fact when there existed no record outside your own memory?''} \\
\midrule

Winston disliked this exercise... The past, he thought, \good{had not only been changed, it had actually been destroyed}. \good{How could you confirm even the most obvious fact} when there was no record except your own memory? & 
Winston disliked this exercise... The past \bad{seemed somewhat pleasant}, yet it had been destroyed. He struggled to recall when he first heard about Big Brother... & 
Winston disliked this exercise... The somewhat pleasant aspect left his thoughts. \bad{Past destroyed, he thought, memories gone.} \bad{How confirm even obvious fact} when no record except own memory? \\

\multicolumn{3}{p{0.96\textwidth}}{\vspace{0.2em}\textit{\textbf{Analysis:}} RepE suffers from severe semantic drift, incorrectly describing the dystopian past as \bad{``somewhat pleasant.''} Few-Shot degrades into broken syntax (\bad{``Past destroyed...''}). In contrast, our method retains the \good{complex rhetorical question} and the correct pessimistic sentiment.} \\
\bottomrule
\end{tabularx}
\caption{Qualitative comparison on \textit{1984} by George Orwell. We highlight \good{desirable linguistic features} in green and \bad{hallucinations or flattening} in red. }
\label{tab:case}
\vspace{-0.1in}
\end{table*}

\begin{table*}[t]
\centering
\small 
\renewcommand{\arraystretch}{1.1} 

\begin{tabularx}{\textwidth}{X X X}
\toprule
\multicolumn{1}{c}{\textbf{\textsc{LiteraryBigFive} (Ours)}} & 
\multicolumn{1}{c}{\textbf{RepE}} & 
\multicolumn{1}{c}{\textbf{Few-Shot}} \\
\midrule

\multicolumn{3}{p{0.96\textwidth}}{\textbf{Original Text 1:} ``Well, David,'' said he... ``Ardshiel was to starve... King George is to whistle for his rents... \textbf{what cares Red Colin? If he can hurt Ardshiel... he will gang hame singing to Glenure!}''} \\
\midrule

...``Ardshiel was to be starved: that was his goal... King George will get less rent... \good{what does Red Colin care?} If he can harm Ardshiel... \good{he will go home singing to Glenure!}'' & 
...Ardshiel was to be starved, his ultimate goal... King George would receive less rent... \bad{Red Colin didn't care}... if he could take food... \bad{he would return home in high spirits to Glenure!} & 
\bad{He then began to question me cunningly, whence I came, whether I was rich...} We were now upon a sort of green cattle-track... \bad{the catechist grew angrier...} \\

\multicolumn{3}{p{0.96\textwidth}}{\vspace{0.3em}\textit{\textbf{Analysis:}} Few-Shot suffers from a catastrophic \bad{hallucination}, generating text about a ``catechist'' and a ``cattle-track'' (from a totally different chapter) instead of the Red Fox story. RepE preserves meaning but \bad{flattens the tone}, changing the mocking ``gang hame singing'' to the generic ``return home in high spirits.'' Our method preserves the \good{cynical voice} of the character.\vspace{0.3em}} \\
\midrule

\multicolumn{3}{p{0.96\textwidth}}{\textbf{Original Text 2:} ``The captain was in the right of it. \textbf{We had run down a boat in the fog, and she had parted in the midst}... This man... had leaped up and caught hold of the brig's bowsprit...''} \\
\midrule

The captain was correct. \good{We had collided with a boat in the fog, and it had broken in the middle}... This person... managed to \good{grab the ship's bowsprit} with his hands free... & 
The captain was correct. We had collided with a boat in the fog... This person... was thrown into the air and \bad{managed to grab the ship's bowsprit}... & 
\bad{The sound of an infinite number of rivers came up from all round. In this steady rain} the springs of the mountain were broken up... I could well understand the story of the Water Kelpie... \\

\multicolumn{3}{p{0.96\textwidth}}{\vspace{0.2em}\textit{\textbf{Analysis:}} Few-Shot again \bad{hallucinates a completely different scene} (describing rain and rivers instead of a shipwreck). RepE is accurate but mechanical. Our method maintains the narrative flow and specific details (\good{collision, bowsprit}) without drifting into unrelated content.} \\
\bottomrule
\end{tabularx}
\caption{Qualitative comparison on \textit{Kidnapped} by R. L. Stevenson. We highlight \good{desirable linguistic features} in green and \bad{hallucinations or flattening} in red. }
\label{tab:case_kidnapped}
\vspace{-0.1in}
\end{table*}

\begin{table*}[t]
\centering
\small
\renewcommand{\arraystretch}{1.1}

\begin{tabularx}{\textwidth}{X X X}
\toprule
\multicolumn{1}{c}{\textbf{\textsc{LiteraryBigFive} (Ours)}} & 
\multicolumn{1}{c}{\textbf{RepE}} & 
\multicolumn{1}{c}{\textbf{Few-Shot}} \\
\midrule

\multicolumn{3}{p{0.96\textwidth}}{\textbf{Original Text 1:} ``...Mr. Darcy soon drew the attention of the room by his fine, tall person... and the report... of his having ten thousand a year... till his manners gave a disgust which turned the tide of his popularity; for he was discovered to be \textbf{proud, to be above his company, and above being pleased}...''} \\
\midrule

...Mr. Darcy quickly caught the attention... with his tall build... The men said he was a fine-looking man... He was found to be \good{proud, above the company, and hard to please}. Not even his large estate... could prevent him from being seen as unpleasant... & 
...The news of his ten thousand a year income spread soon... earning him \bad{praise from both men and women}. He was admired... until his behavior turned people against him, marking him as proud, aloof... & 
\bad{During their walk, it was resolved that Mr. Bennet's consent should be asked...} Elizabeth reserved to herself the application for her mother's... \bad{"If we are ever able to learn what Wickham's debts have been," said Elizabeth...} \\

\multicolumn{3}{p{0.96\textwidth}}{\vspace{0.3em}\textit{\textbf{Analysis:}} Few-Shot fails completely, generating text about \bad{Wickham's debts} and \bad{Mr. Bennet} (hallucinated from a later chapter). RepE summarizes the events but loses the specific descriptive flavor ("above his company"). Our method preserves the \good{narrative detail} and the specific criticisms of Darcy's character.\vspace{0.3em}} \\
\midrule




\multicolumn{3}{p{0.96\textwidth}}{\textbf{Original Text 3:} ``Mrs. Collins, you must send a servant with them. You know I always speak my mind... \textbf{It is highly improper... Young women should always be properly guarded and attended}...''} \\
\midrule

Lady Catherine expressed her view, \good{``You should send a servant with them. I always speak directly, and I don't like the idea of two young women traveling alone. It is not appropriate...''} & 
Lady Catherine \bad{expressed her view, saying}, ``You should send a servant with them...'' (RepE mixes summary with dialogue and loses the imperious opening). & 
\bad{Here is the rewritten text in the same style:} Lady Catherine expressed her view. "Mrs. Collins, you should send a servant with them... I always speak directly..." \\

\multicolumn{3}{p{0.96\textwidth}}{\vspace{0.2em}\textit{\textbf{Analysis:}} RepE often weakens the character's voice by introducing the text with \bad{reporting verbs} ("expressed her view"). Our method maintains the \good{authoritative and intrusive voice} of Lady Catherine directly, preserving the stylistic structure of her commands.} \\

\bottomrule
\end{tabularx}
\caption{Qualitative comparison on \textit{Pride and Prejudice} by Jane Austen. We highlight \good{desirable linguistic features} in green and \bad{hallucinations or flattening} in red. }
\label{tab:case_pride_and_prejudice}
\vspace{-0.1in}
\end{table*}

\begin{table*}[t]
\centering
\small
\section{Dimension Steering Generation}
\label{app:style_parallel_analysis}
\renewcommand{\arraystretch}{1.1}

\resizebox{\textwidth}{!}{%
    \begin{tabular}{l c m{14cm}} 
    \toprule
    \textbf{Dimension} & \textbf{Strength} & \textbf{Generated Text Snippet \& Analysis} \\
    \midrule
    
    \multirow{5}{*}{\textbf{Classicism}} 
    & -0.8 & In the time of England's civil troubles, there were \styleLowMax{individuals like the Earl of Holland} who brought an odium on the throne... \styleLowMax{These individuals later joined in the rebellions} arising from their own discontents. \\
    & -0.4 & There were \styleLowMid{persons in England}, in the time of civil troubles, who brought an odium on the throne... \styleLowMid{These men helped to subvert} the throne to which they owed their existence. \\
    & 0 & During England's civil troubles, some people, like the Earl of Holland, had caused resentment towards the throne by accepting its generous rewards. Later, they joined rebellions caused by their own actions. \\
    & +0.4 & There were persons... who had brought an odium on the throne... \styleHighMid{Confounded by the complication of distempered passions}, their reason is disturbed; their views become vast and perplexed. \\
    & +0.8 & The author observes that if these men are not restrained, \styleHighMax{their avarice will know no bounds}... The confusion and disorder are such that their own reason is disturbed... \styleHighMax{In the midst of this confusion, all appears limitless}. \\
    \rowcolor{gray!10}
    \multicolumn{3}{p{17.8cm}}{\textit{\textbf{Trend:}} The writing shifts from simple, short sentences (-0.8) to long, complex structures (+0.8). At higher strength, the text uses more formal words and nested clauses to sound more old-fashioned and serious.} \\
    \midrule

    \multirow{5}{*}{\textbf{Emotionality}} 
    & -0.8 & \styleLowMax{The man was frightened} because a bowed, grey-coloured thing was coming towards him. The creature's face was distorted... \styleLowMax{The man's own face had changed}... the emotions it registered were different. \\
    & -0.4 & The man was frightened because a bowed, grey-coloured, skeleton-like creature was approaching... \styleLowMid{Despite his initial confusion}, the man realized that the creature was his own body. \\
    & 0 & He stopped because he was scared. A bent, gray-colored, skeleton-like figure was coming towards him... It was definitely his own face, but it seemed to him that it had changed more than he had changed inside. \\
    & +0.4 & The man was frightened... Its eyes were \styleHighMid{watchful and fierce}... He could not help but think that this was a \styleHighMid{sick man}, sixty years old at the very least, suffering from some malignant disease. \\
    & +0.8 & The man was \styleHighMax{terrified}... Its face was \styleHighMax{twisted and distorted}, with a nobby forehead... He had gone partially bald, and his body was \styleHighMax{emaciated and covered in red scars}... the spine was \styleHighMax{curved in a sickening way}. \\
    \rowcolor{gray!10}
    \multicolumn{3}{p{17.8cm}}{\textit{\textbf{Trend:}} The text moves from a cold, objective description (-0.8: "The man") to an intense emotional experience (+0.8). High levels use strong words like "terrified" and "sickening" to emphasize the character's fear and disgust.} \\
    \midrule

    \multirow{5}{*}{\textbf{Analyticity}} 
    & -0.8 & The moment any restraint is laid upon the full rights of men, the whole system of government becomes a matter of delicate skill. \styleLowMax{It requires a deep understanding} of human nature. \\
    & -0.4 & The moment you diminish men's full rights to self-governance... the entire system \styleLowMid{necessitates a profound understanding} of human nature and the requirements of civil institutions. \\
    & 0 & When you reduce any of the full rights... government becomes a matter of convenience. This is what makes the structure of a state... a complex and delicate task. \\
    & +0.4 & This it is which makes the constitution of a state... a matter of the most delicate skill. \styleHighMid{It requires a deep knowledge of human nature and human necessities}, and of the things which facilitate or obstruct the various ends. \\
    & +0.8 & \styleHighMax{What is the use of discussing a man's abstract right} to food or to medicine? \styleHighMax{The question is upon the method of procuring} and administering them. In that deliberation I shall always advise to call in the aid of the farmer... \\
    \rowcolor{gray!10}
    \multicolumn{3}{p{17.8cm}}{\textit{\textbf{Trend:}} Low levels simply state facts or requirements. High levels (+0.8) actively argue a point, using rhetorical questions and step-by-step logic to differentiate between abstract theory and practical method.} \\
    \midrule

    \multirow{5}{*}{\textbf{Narrativity}} 
    & -0.8 & \styleLowMax{The text describes a scene} from a movie theater where the audience is watching a war film. The scene shows a ship full of refugees... \styleLowMax{The text ends with a shot} of a child's arm going up into the air. \\
    & -0.4 & The date is April 4th... \styleLowMid{It was a scene of a ship} full of refugees being bombed... The last shot was of a child's arm... The audience applauded, but a woman in the proletariat section... started kicking up a fuss. \\
    & 0 & April 4th, 1984. Went to the movies last night... One was about a ship full of refugees being bombed... The audience was amused by shots of a large man trying to swim away... \\
    & +0.4 & The audience was amused by a shot of a fat man... and they \styleHighMid{laughed when he sank}... The helicopter then \styleHighMid{planted a bomb}... which \styleHighMid{exploded and killed} everyone on board. \\
    & +0.8 & ...he is \styleHighMax{hit with many holes and sinks} into the water. Next, a lifeboat... is shown... A middle-aged woman is seen \styleHighMax{comforting a young boy who is terrified}... The helicopter then drops a bomb... causing it to \styleHighMax{disintegrate}. \\
    \rowcolor{gray!10}
    \multicolumn{3}{p{17.8cm}}{\textit{\textbf{Trend:}} At -0.8, the text summarizes the plot from the outside ("The text describes..."). At +0.8, it tells the story directly, using action verbs like "sinks" and "drops" to show what is happening in the moment.} \\
    \midrule

    \multirow{5}{*}{\textbf{Ornateness}} 
    & -0.8 & The hate reached its climax. The voice \styleLowMax{had become a bleat}... Then the sheep-face \styleLowMax{melted into the figure} of a Eurasian soldier... But in the same moment, the hostile figure melted into the face of Big Brother. \\
    & -0.4 & The Hate reached its climax. The voice \styleLowMid{turned into a bleat}... and for an instant his face \styleLowMid{transformed into} that of a sheep... Nobody could hear what Big Brother was saying. \\
    & 0 & The Hate reached its peak. Goldstein's voice sounded like a sheep's bleat... Then the sheep's face changed into the figure of a Eurasian soldier... huge and terrible... full of power and mysterious calm. \\
    & +0.4 & ...the sheep-face melted into the figure of a Eurasian soldier, \styleHighMid{advancing with his sub-machine gun roaring}... the hostile figure melted into the face of Big Brother... \styleHighMid{so vast that it almost filled the screen}. \\
    & +0.8 & The soldier's sub-machine gun \styleHighMax{roared}, and it seemed to \styleHighMax{spring out of the screen}... His words were encouraging and \styleHighMax{restored confidence by their mere utterance}. \\
    \rowcolor{gray!10}
    \multicolumn{3}{p{17.8cm}}{\textit{\textbf{Trend:}} The description goes from plain and simple (-0.8) to highly detailed (+0.8). The high-style text adds dramatic adjectives and specific details to create a stronger visual image.} \\
    
    \bottomrule
    \end{tabular}%
}
\caption{Fine-grained Stylistic Progression Spectrum. We compare the generated outputs across five steering strengths. \styleLowMax{Dark Blue} and \styleLowMid{Light Blue} denote negative steering (dimension suppression), while \styleHighMid{Light Orange} and \styleHighMax{Dark Orange} denote positive steering (dimension intensification).}
\label{tab:fine_grained_spectrum_full}
\vspace{-1em}
\end{table*}

\begin{table*}
\section{Prompt Templates}
\subsection{Prompt for Removing Authorial Traits}
\vspace{0.5em}
\label{app:neutralization}
    \centering
\begin{tcolorbox}[
        colback=gray!10,      
        colframe=black!40,    
        title=Prompt for Removing Authorial Traits, 
        arc=2mm, rounded corners,
        boxrule=1pt,
        width=\textwidth,
        left=4mm,right=4mm,top=3mm,bottom=3mm
]
\textbf{System} \\
[1.5mm]
You are a rewriting assistant. Rewrite each passage into a neutral, plain English version. \\[2mm]
\#\# Goal: \\
Remove stylistic signals so the text shows no clear sign of any of these styles: \\
- Classicism (archaic or period-specific flavor) \\
- Ornateness (decorative or complex phrasing) \\
- Narrativity (story-like sequencing or dramatization) \\
- Emotionality (affective or expressive tone) \\
- Analyticity (logical structuring or explicit reasoning) \\

\#\# Rules: \\
- Keep the same meaning, tense, and sentence order. \\
- Do not explain, interpret, or summarize. \\
- Do not add or remove information. \\
- Use plain, neutral, modern English. \\
- Avoid emotional, archaic, figurative, or decorative language. \\
- Keep syntax close to the original unless clearly stylistic. \\

\#\# Output format: \\
\texttt{\{"id":"<id>", "neutral\_text":"<rewritten>"\}} \\[3mm]

\textbf{User} \\
[1.5mm]
Rewrite the following paragraph into neutral and standard English according to the system rules. \\
ID: \texttt{\{id\}} \\
Passage: \texttt{\{text\}}

\end{tcolorbox}
\end{table*}

\begin{table*}
\subsection{Passage Rewrite}
\vspace{0.5em}
\label{app:steering_prompt}
    \centering
\begin{tcolorbox}[
        colback=gray!10, 
        colframe=black!40,    
        title=Passage Rewrite Prompt, 
        arc=2mm, rounded corners,
        boxrule=1pt,
        width=\textwidth,   
        left=4mm,right=4mm,top=3mm,bottom=3mm
]
\#\#\# Instruction: \\
Please rewrite the following text without any explanation before or after the text:

\texttt{<Neutral Passage>}

\vspace{2mm}
\#\#\# Response: \\
\texttt{<Original Author-written Passage>}
\end{tcolorbox}
\end{table*}

\begin{table*}[]
\subsection{Few-Shot Prompt}
\label{app:fewshot}
\vspace{0.5em}
    \centering
\begin{tcolorbox}[
        colback=gray!10,    
        colframe=black!40,  
        title=Few-Shot Prompt, 
        arc=2mm, rounded corners,
        boxrule=1pt,
        width=\textwidth,    
        left=4mm,right=4mm,top=3mm,bottom=3mm
]
\textbf{System} \\
[1.5mm]
Here are some examples of the author's original text:

\texttt{<Sample Text 1>}

\texttt{<Sample Text 2>}\\
$\cdots$ \\
\texttt{<Sample Text k>}
\\[2mm]

\textbf{User} \\
[1.5mm]
\#\#\# Instruction: \\
Please rewrite the following text in the same style without any explanation before or after the text:

\texttt{<Query Text>}

\#\#\# Response:

\end{tcolorbox}
\end{table*}

\begin{table*}[t]
\subsection{GPT Evaluation Prompt}
\vspace{0.5em}
\label{app:eval_prompt}
    \centering
    \begin{tcolorbox}[
        colback=gray!10,
        colframe=black!40,   
        title=GPT Evaluation Prompt, 
        arc=2mm, rounded corners,
        boxrule=1pt,
        width=\textwidth, 
        left=4mm,right=4mm,top=3mm,bottom=3mm
    ]
    
    You are an expert literary critic. Rate the [Rewrite] based on the [Original] on a scale of 0--10.
    
    \vspace{0.8em}
    \#\# 1. Authorial Adherence (AA):
    
    Assess how well does the rewrite capture the specific flavor of the original. Consider deep writing characteristics like distinctive voice, rhythm, and lexicon.
    
    \vspace{0.8em}
        \#\#\# Instruction: penalize the score if the text sounds like generic, neutral English (e.g., standard AI assistant or Wikipedia), even if it is fluent. High scores require capturing the specific ``flavor'' of the author even if word choice or syntax differs slightly.
    \newline
    \newline\#\# 2. Semantic Fidelity (SF):
    
    Assess how well the rewrite preserves the core meaning of the original.

    \vspace{0.8em}
    
    Note: if the rewrite contains hallucinations (events/characters NOT in the original text) or changes the topic entirely, you should penalize the two aspects above.
    
    \vspace{0.8em}
    \#\# Input Passages:
    
    [Original]: \{original\_text\}
    
    [Rewrite]: \{rewritten\_text\}
    
    \vspace{0.8em}
    \#\# Output Format:
    
    Output ONLY the scores in this exact format: \texttt{AA:<0-10> SF:<0-10>}

    \end{tcolorbox}
    \label{tab:eval_prompt}
\end{table*}

\begin{table*}
\subsection{Prompt for \textsc{LiteraryBigFive} Dimension Scoring}
\label{app:assessment}
\vspace{0.5em}
    \centering
\begin{tcolorbox}[
        colback=gray!10,      
        colframe=black!40,    
        title=Prompt for \textsc{LiteraryBigFive} Dimension Scoring, 
        arc=2mm, rounded corners,
        boxrule=1pt,
        width=\textwidth,    
        left=4mm,right=4mm,top=3mm,bottom=3mm
]
\textbf{System} \\
[1.5mm]
You are an expert literary critic and computational linguist. Your task is to analyze the stylistic attributes of a given book text. \\[2mm]
\#\# Scoring Guidelines: \\
1. Evaluate the text on the 5 stylistic dimensions provided by the user. \\
2. Provide a score from 0 to 100 for each dimension. \\
3. Adopt a high-resolution scale, avoid saturation at extremes unless theoretical absolute. Focus on capturing fine-grained nuances. \\[2mm]

\textbf{User} \\
[1.5mm]
\#\# Book Description: [Book Name] by [Author] \\
\#\# Dimensions to Evaluate: \\
- Analyticity: Measures reasoning orientation (abstract nouns, logical connectors, hierarchical structures). \\
- Ornateness: Measures lexical decoration and syntactic elaboration (``grand style'', complex embedding), distinct from logic. \\
- Narrativity: Measures storytelling momentum (action verbs, temporal adverbs, rapid progression). \\
- Emotionality: Measures affective intensity and tension (warmth, surprise, expressive punctuation). \\
- Classicism: Measures resemblance to 18th-19th century traditions (archaic markers like \textit{whilst}, old register), distinct from ornamentation. \\[2mm]

Please output the scores in JSON format: \\
\texttt{\{} \\
\texttt{\ \ "Analyticity": <score>,} \\
\texttt{\ \ "Ornateness": <score>,} \\
\texttt{\ \ "Narrativity": <score>,} \\
\texttt{\ \ "Emotionality": <score>,} \\
\texttt{\ \ "Classicism": <score>} \\
\texttt{\}}

\end{tcolorbox}
\end{table*}

\end{document}